\documentclass [twocolumn,final] {svjour3}
\usepackage{setspace}
\usepackage{hyperref}
\usepackage{color}
\usepackage{amssymb}

\usepackage{amsmath, amsthm}
\usepackage{mathrsfs}
\usepackage[linesnumbered,ruled,vlined]{algorithm2e} 
\usepackage{enumitem}  
\usepackage{pgfplots}
\usepackage{tikz}
\usepackage{graphicx}
\usepackage{threeparttable}
\usepackage{makecell}
\usepackage{multirow}
\usepackage{mathtools}
\usepackage[caption=false,font=footnotesize]{subfig}
\usepackage{textcomp}
\usepackage{gensymb}
\usepackage{esvect}
\usepackage{siunitx}
\usepackage{pifont}
\usetikzlibrary{arrows,calc,patterns}
\usepackage{wrapfig}

\theoremstyle{definition}

\newtheorem{defi}{Definition}[subsection]
\newtheorem{rem}{Remark}[subsection]

\usepackage{chngcntr}
\counterwithin{figure}{section}
\counterwithin{table}{section}
\numberwithin{equation}{section}

\newcommand{\xmark}{\text{\ding{55}}}
\newcommand{\cmark}{\ding{51}}

\DeclareMathOperator*{\argmax}{arg\,max}
\pgfplotsset{compat=1.12}

\title{A Two-Stage Method for Text Line Detection in Historical Documents}

\author{Tobias Gr\"uning \and Gundram Leifert \and Tobias Strau\ss \and Johannes Michael \and Roger Labahn}
\institute{Tobias Gr\"uning \and Tobias Strau\ss
	\at Planet AI, Warnowufer 60, 18057 Rostock, Germany \\\email{tobias.gruening@planet.de}
	\and
	Gundram Leifert \and Johannes Michael \and Roger Labahn
	\at Computational Intelligence Technology Lab, Department of Mathematics, University of Rostock, 18057 Rostock, Germany
}
\date{Received: date / Accepted: date}

\begin{document}\sloppy
\maketitle

\begin{abstract}
	This work presents a two-stage text line detection method for historical documents. Each detected text line is represented by its baseline. 
	In a first stage, a deep neural network called ARU-Net labels pixels to belong to one of the three classes:  baseline, separator or other. 
	The separator class marks beginning and end of each text line. 
	The ARU-Net is trainable from scratch with manageably few manually annotated example images (less than $50$). 
	This is achieved by utilizing data augmentation strategies. 
	The network predictions are used as input for the second stage which performs a bottom-up clustering to build baselines. 
	The developed method is capable of handling complex layouts as well as curved and arbitrarily oriented text lines. 
	It substantially outperforms current state-of-the-art approaches. For example,
	for the complex track of the cBAD: ICDAR2017 Competition on Baseline Detection the F-value is increased from $0.859$ to $0.922$. 
	The framework to train and run the ARU-Net is open source.
\end{abstract}

\section{Introduction}
\setcounter{subsection}{1}
Accessibility of the valuable cultural heritage of historical documents is an important concern of archives, libraries as well as certain companies, e.g., those specialized in 
genealogy. After years of digitization at an industrial scale to protect and preserve these valuable goods, millions over millions of scanned pages are
stored at servers all over the world 
\cite{isaac2012}. The generic next step is to make the enormous amount of content of these document images accessible and enable humanists, historians, genealogists as well as
ordinary people to efficiently work with these documents. Besides the cost- and time-consuming process of manually annotating volumes \cite{Causer2012}, it is subject to current research and scientific discussion 
how to automate this process \cite{sanchez2013transcriptorium}.

Since 2009, tremendous progress in the field of Automated Text Recognition\footnote{Optical Character Recognition + Handwritten Text Recognition} (ATR) \cite{Graves2008,leifert16} as well as Keyword Spotting (KWS) \cite{Puigcerver2014,strauss2016citlab,Strauss2016} was achieved.
The performance of state-of-the-art systems reaches character error rates below $10 \%$ for ATR \cite{Sanchez2016} and mean average precisions above $0.9$ for KWS \cite{Pratikakis2016} for complex handwritten documents. 
Although efforts are made to develop systems working sole\-ly on the rough input image without any a-priori segmentation \cite{Rusinol2015,Bluche2016,Konidaris2016},
the best performing recognition systems -- with reference to recently hosted competitions -- rely on segmented words or text lines as input. Entirely segmentation-free approaches suffer either from an enormous training/inference time and/or, up to now, 
did not demonstrate its applicability with competitive quality on challenging datasets \cite{Pratikakis2016}.
Hence, a workflow which involves a text line extraction followed by the transformation of pixel information into textual information (ATR/KWS) is the widely used standard. 
This work deals with the first step of the information retrieval pipeline, namely the text line extraction. This is a mandatory step since errors directly effect the performance of the overall information retrieval process. 
The text line extraction is still unsolved to a certain extent for historical documents due to difficulties such as physical degradations (e.g., bleed-through, faded away characters, heterogeneous stroke intensity), image capture conditions (e.g., scan curve, illumination issues),
complex layouts (e.g., structured documents, marginalia, multi-column layouts, varying font sizes), arbitrary orientations and curved text lines.

The results achieved by state-of-the-art approaches are not satisfying \cite{Murdock2015}, especially if dealing with heterogeneous data.
Therefore, this work focuses on the extraction of text lines in arbitrary historical documents.
Since different ATR/KWS systems necessitate different text line representations, e.g., bounding boxes \cite{Sudholt2016}, x-height areas \cite{renton2018fully} or more precise polygonal representations following all ascenders and descenders \cite{strauss2016citlab}, 
there is not the \textbf{one} correct text line representation.
Therefore we limit ourselves towards the text line detection task by representing each text line by its baseline. 
The detected baselines allow for an extraction of the text lines in an appropriate -- with respect to the following method -- way. The problem of extracting a text line given its baseline has to handle problems like touching or overlapping components and can be tackled by applying, e.g.,
histogram approaches to estimate the x-height \cite{renton2018fully} or by utilizing Dynamic Programming to calculate separating seams \cite{Arvanitopoulos2014} However, this is not within the scope of this work.

Besides the classical image processing based approaches, deep learning based methods became omnipresent in the document analysis community within the last years. 
Such techniques were recently used to solve several different problems such as binarization \cite{Vo2017}, page boundary extraction \cite{byu2017b}, page segmentation \cite{chen2017} or text line detection \cite{renton2018fully}.
The presented work to our knowledge is the first which uses a two-stage method, combining deep learning strategies and state-of-the-art image processing based techniques.
We propose an extension of the U-Net \cite{Ronneberger2015}, the so-called ARU-Net. The fully convolutional U-Net is extended by incorporating residual blocks \cite{He2016} to increase its representative power.
Furthermore, a spatial attention mechanism is developed which allows the ARU-Net to focus on image content at different positions and scales.
The network is designed to processes the entire, arbitrarily-sized image at once to take account of all spatial context.
The ARU-Net is universal in a way, that it could be used to tackle any pixel labeling task. 
In this work, it is trained in a fully supervised fashion to classify each pixel to belong to one of the following classes: baseline, separator or other. 
The separator class is introduced to explicitly predict beginning and end of each text line and not just rely on the information implicitly given by the baseline class. 
This is advantageous for text lines which are close together but have to be separated, e.g., those belonging to different columns. 
The network output serves as input for an image processing based bottom-up clustering approach. This approach utilizes so-called states of superpixels \cite{Ryu2014}, which encode local text orientation and interline distances.
This second stage allows for an error correction of the network output by incorporating domain knowledge based on assumptions, which hold for text lines in general, see Sec.~\ref{sssec:sc}.
Additionally, it is easily possible to incorporate the separator information, which allows for a handling of documents with complex layouts, e.g., 
images containing tables or marginalia. 

Each method relying on supervised deep learning and therefore relying on training data can suffer from the need of an enormous amount of labeled training data. 
We demonstrate that the presented approach achieves high quality results on the Bozen dataset \cite{bozen} with less
than $50$ full-page training samples by using data augmentation strategies. Along with an annotating effort of just a few minutes per page the adaptation of the proposed method is easy and cheap.
We demonstrate the applicability of the proposed method for images with arbitrarily oriented as well as curved text lines by achieving nearly as good results as for straight $0\degree$ oriented text lines.
Finally, we show that the presented approach outperforms state-of-the-art methods on three different datasets. A relative F-value \cite{gruning2017read} error (the gap to 1.0) reduction of at least $24\%$ is achieved for the cBAD dataset \cite{diem_markus_2017_257972}. This dataset is 
composed of $2036$ historical images with annotated baselines of nine different archives and libraries from all over Europe and is therefore -- in the opinion of the authors -- the most representative and heterogeneous freely available dataset.
Especially, for the complex track, which contains mostly documents with complex layouts, the average F-value is increased from $0.859$ to $0.922$. 

The main contributions of this work are:
\begin{itemize}
 \item introduction of a newly designed deep neural network (ARU-Net) for pixel labeling along with a meaningful parametrization -- the ARU-Net and its training framework are open source\footnote{https://github.com/TobiasGruening/ARU-Net},
 \item introduction of the new concept of learned separators to handle complex layouts instead of an a-priori page segmentation or white-/blackrun calculation
 \item introduction of a state-of-the-art two-stage workflow which combines state-of-the-art deep learning and image processing techniques -- the entire workflow is freely usable via the Transkribus platform\footnote{https://transkribus.eu}.
\end{itemize}

 
\section{Related Work}
\label{sec:rel}
A comprehensive survey of approaches for text line extraction in historical documents is given in \cite{Likforman-Sulem2007} and \cite{Eskenazi2017}. 
In this section, we will focus on approaches relevant for this work.

In \cite{Arvanitopoulos2014,Nicolaou2009,Saabni2014}, the principle of Dynamic Programming is utilized to calculate cost optimal 
paths passing the image from left to right to separate different text lines from each other. These methods basically differ in the way the images are pre-processed and in the definition of the cost function. Garz et al. \cite{Garz2012} propose 
a method based on clustering of interest points (this is just another name for what we call superpixel). Using a standard clustering technique, interest points in an area which exceeds a certain density are clustered to form word clusters. Word clusters are separated to sub-word 
segments and these are finally grouped to build text lines. 
Ryu et al. \cite{Ryu2014} propose an algorithm which uses certain characteristics (so-called states) of extracted connected components to assign costs to certain clustering results. 
These states encode local text orientation and interline distances and are introduced in Def.~\ref{def:state}. Subsequently using
four different operations (merge, split, merge-split, merge-merge-split) on an initial coarse clustering, the costs are minimized to obtain an optimal clustering, which leads to the final text line segmentation.
Ahn et al. \cite{Ahn2017} improve this approach by the introduction of a newly developed binarization method and an improved clustering process.
Gr\"uning et al. \cite{gr2017} extended the approach of Ryu et al. so that it is applicable for more general superpixels with a newly introduced clustering procedure which does not rely on a coarse initial clustering.
Besides these ``classical'' approaches, which are based on image processing techniques, methods based on machine learning gained importance within the last two years.
Moysset et al. \cite{Moysset2015} propose a method based on a recurrent neural network.
The network is trained given only the number of lines in the image utilizing Connectionist Temporal Classification which was introduced to train networks for handwriting text recognition and allows for ground truth data without any alignment. The trained neural network predicts confidences for the vertical coordinates of the image to belong either to the classes line or interline. Further post-processing
of the neural network output is performed to detect the text lines. In follow-up works, they formulated the problem as a regression problem \cite{moysset2018learning}. The recurrent neural network directly predicts bounding boxes as well as the start of each text line, respectively.
Besides this regression based approach, classification based approaches were proposed most recently. In contrast to the approach of Moysset et al., these methods perform a pixel labeling to classify each image pixel (instead of classifying rows of pixels, only). For instance,
Renton et al. \cite{renton2018fully} propose a fully convolutional network (FCN) based on dilated (or atrous) convolutions to classify pixels as text line main body or not. The classification results are utilized to extract the text line information. 
These techniques are currently very popular, e.g., four of the five participants of the cBAD: ICDAR2017 Competition on Baseline Detection \cite{diem2017} use methods relying on FCNs. E.g., the methods presented in \cite{renton2018fully,quiros2018multi,fink2018baseline,oliveira2018dhsegment} tackle the problem of baseline detection with fully convolutional neural networks. However, these methods either rely on a patch-wise processing of the input image (which results in a cumbersome reconstruction of the baseline hypothesis for the entire image), on massive pre-trained encoder structures (which significantly increases the number of model parameter and slows down the system) or on elementary post processing steps (which deteriorates the system's performance). The presented method overcomes this limitations and shows its superiority on the challenging cBAD dataset.
\section{Methodology}
\label{sec:met}
\newlength{\mywidth}
\setlength{\mywidth}{100pt}
\tikzstyle{copy} = [color=black!80,thick,->]
\newcommand{\drawWF}[2]
{
    \tikzstyle{fancytitle} =[fill=white, text=black]
    \begin{tikzpicture}
    
      \node (pic1) at (0,0) {\includegraphics[width=\mywidth]{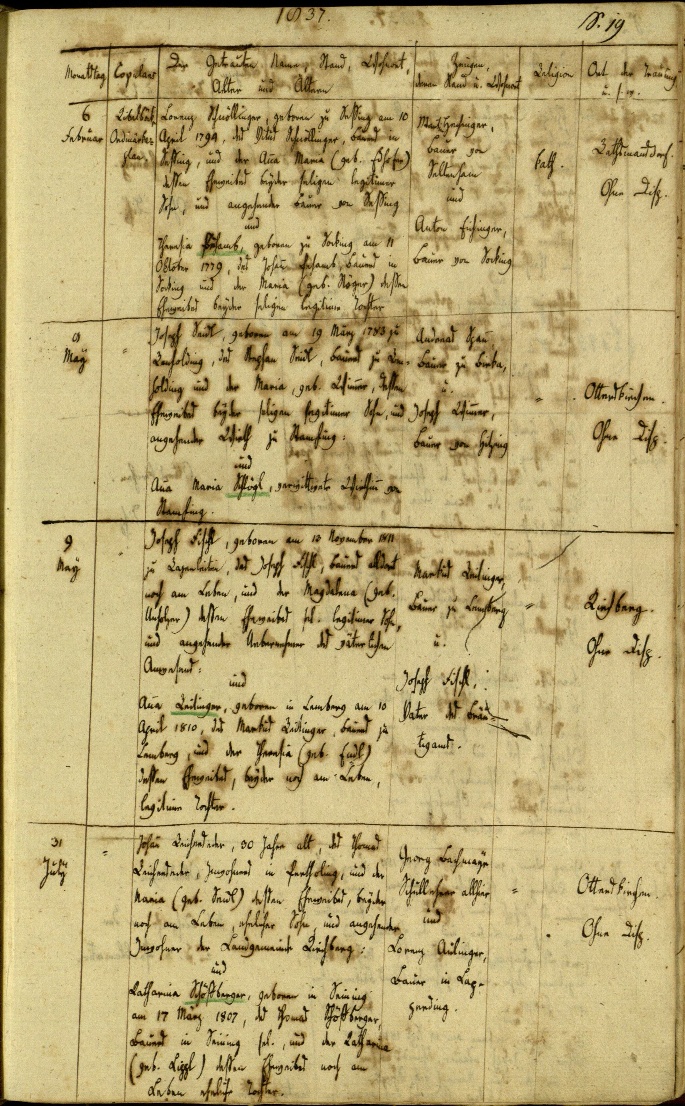}};
      \node [draw, dotted, thick, rectangle, rounded corners,minimum width=\mywidth+10pt,minimum height=1.85*\mywidth] (box1) at (0,0){};
      \node[fancytitle, right=5pt] at (box1.south west) {\scriptsize Input};

      \node [draw, rectangle, rounded corners,minimum width=1.25*\mywidth,rotate=90, minimum height=#2pt] (net) at (0.5\mywidth+0.5*#2+2*#1pt,0){\scriptsize ARU-Net};
      \node (pic2) at (0.725*\mywidth+#2+3*#1pt ,0.4*\mywidth) {\includegraphics[width=0.45\mywidth]{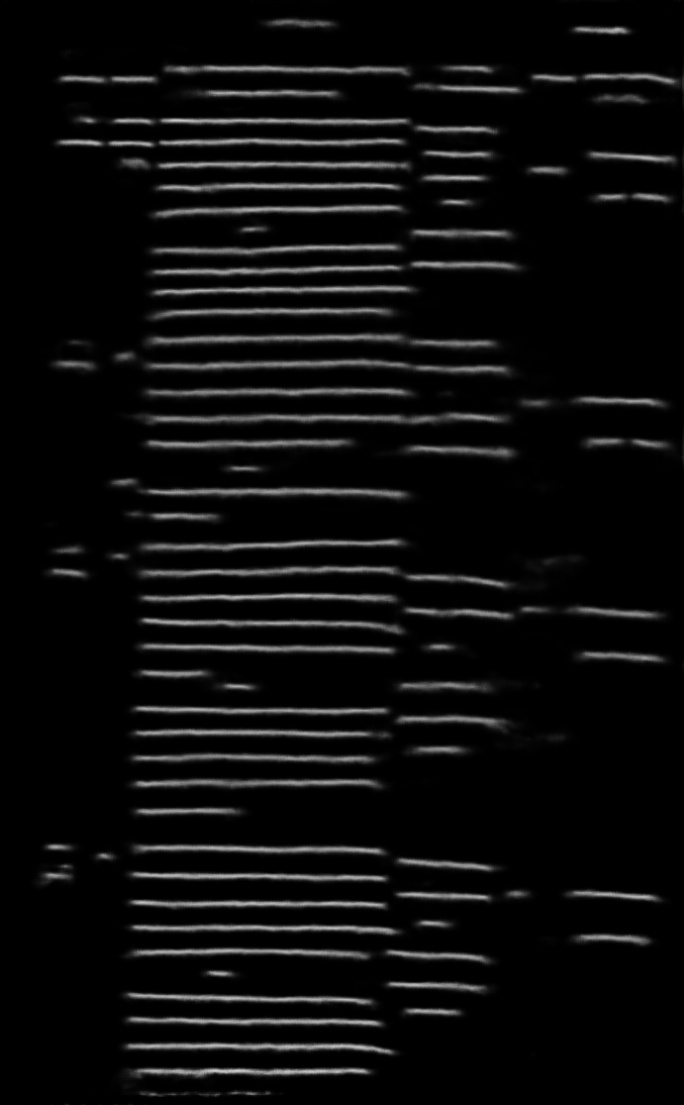}};
      \node (pic3) at (0.725*\mywidth+#2+3*#1pt,-0.4*\mywidth) {\includegraphics[width=0.45\mywidth]{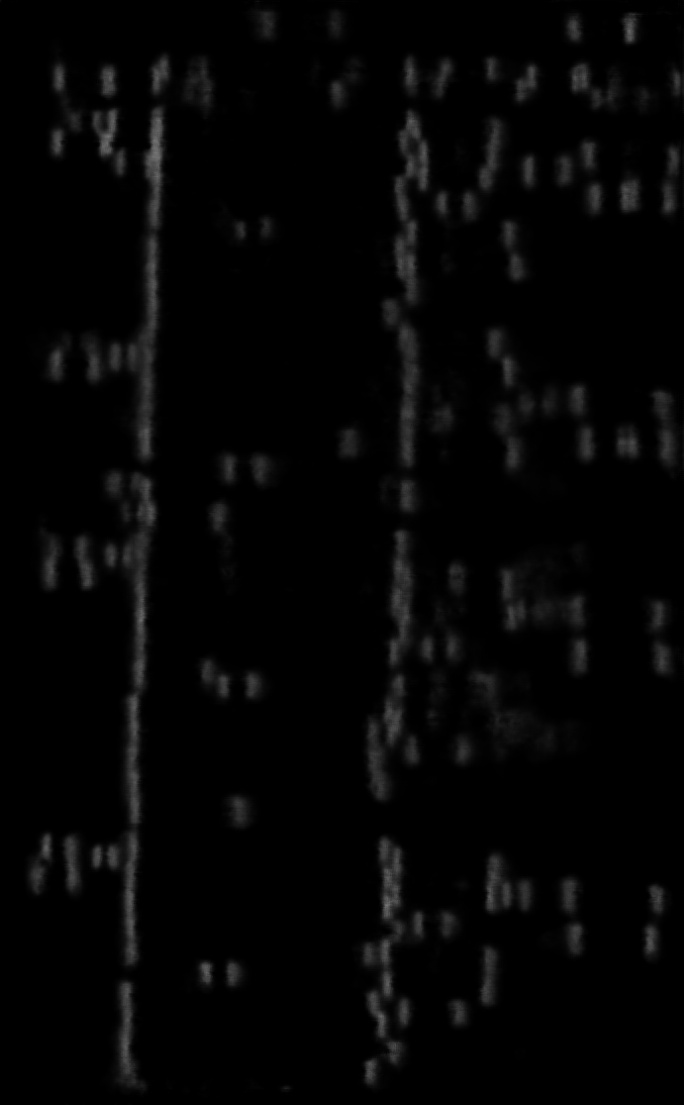}};
      \node [draw, dotted, thick, rectangle, rounded corners,minimum width=0.45\mywidth+#1+#2+10pt,minimum height=1.85*\mywidth] (box2) at (0.725*\mywidth+0.5*#2+2.5*#1pt,0){};
      \node[fancytitle, right=5pt] at (box2.south west) {\scriptsize Stage I};
      
      \node [draw, rectangle, rounded corners,minimum width=1.25*\mywidth,rotate=90, minimum height=#2pt] (spc) at (0.95*\mywidth+1.5*#2+5*#1pt,0){\scriptsize Superpixel Calculation};
      \node [draw, rectangle, rounded corners,minimum width=1.25*\mywidth,rotate=90, minimum height=#2pt] (ste) at (0.95*\mywidth+2.5*#2+6*#1pt,0){\scriptsize State Estimation};
      \node [draw, rectangle, rounded corners,minimum width=1.25*\mywidth,rotate=90, minimum height=#2pt] (cl) at (0.95*\mywidth+3.5*#2+7*#1pt,0){\scriptsize Superpixel Clustering};
      \node [draw, dotted, thick, rectangle, rounded corners,minimum width=3*#2+2*#1+10pt,minimum height=1.85*\mywidth] (box3) at (0.95*\mywidth+2.5*#2+6*#1pt,0){};
      \node[fancytitle, right=5pt] at (box3.south west) {\scriptsize Stage II};

      \node (pic4) at (1.45*\mywidth+4.0*#2+9*#1pt,0) {\includegraphics[width=\mywidth]{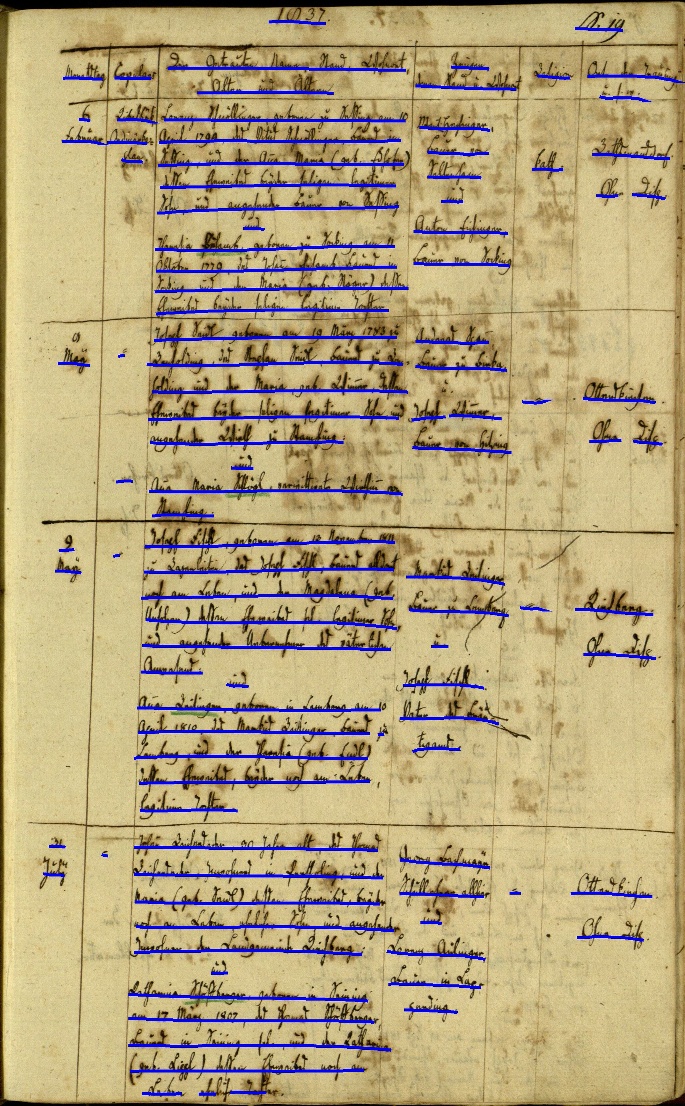}};
      \node [draw, dotted, thick, rectangle, rounded corners,minimum width=\mywidth+10pt,minimum height=1.85*\mywidth] (box4) at (1.45*\mywidth+4.0*#2+9*#1pt,0){};
      \node[fancytitle, right=5pt] at (box4.south west) {\scriptsize Output};

      \draw [copy] (0.5\mywidth, 0)--(net);
      \draw [copy] (net)--(0.5\mywidth+#2+2.5*#1pt,0)--(0.5\mywidth+#2+2.5*#1pt,0.4*\mywidth)--(0.5*\mywidth+#2+3*#1pt ,0.4*\mywidth);
      \draw [copy] (net)--(0.5\mywidth+#2+2.5*#1pt,0)--(0.5\mywidth+#2+2.5*#1pt,-0.4*\mywidth)--(0.5*\mywidth+#2+3*#1pt ,-0.4*\mywidth);

      \draw [copy] (0.95*\mywidth+#2+3*#1pt ,0.4*\mywidth)--(0.95*\mywidth+#2+4*#1pt ,0.4*\mywidth)--(0.95*\mywidth+#2+4*#1pt ,0)--(spc);
      \draw [copy] (0.95*\mywidth+#2+3*#1pt ,-0.4*\mywidth)--(0.95*\mywidth+#2+4*#1pt ,-0.4*\mywidth)--(0.95*\mywidth+#2+4*#1pt ,0)--(spc);

      \draw [copy] (spc)--(ste);
      \draw [copy] (ste)--(cl);
      \draw [copy] (cl)--(0.95*\mywidth+4.0*#2+9*#1pt,0);
		
    \end{tikzpicture}
}
\begin{figure*}[ht]
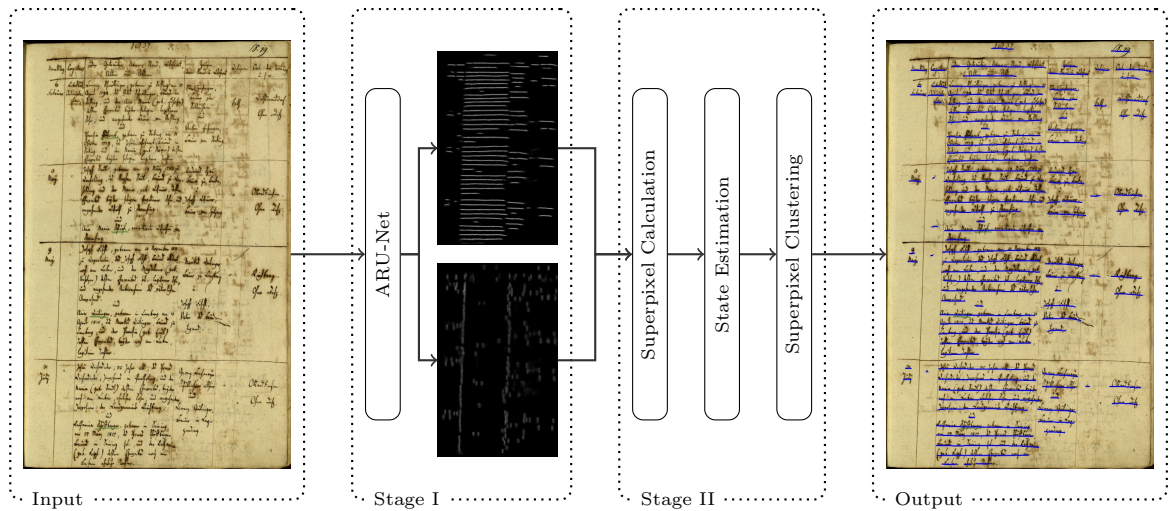

	\centering
	\drawWF{14}{13}
	\caption{\textbf{Two-stage workflow to detect baselines --} The first stage utilizes a deep hierarchical neural network to perform a pixel labeling. 
	The result of Stage I is the input for an image processing based method in Stage II. This method clusters superpixel to build baselines. The image is sampled from the cBad complex test set \cite{gruning2017read}.}
	\label{fig:wf}
\end{figure*}
In this section, we introduce the two-stage method for baseline detection, see Fig.~\ref{fig:wf}. The first stage relies on a deep neural network -- the ARU-Net -- and performs a pixel labeling. 
The pixel labeling can be seen as some kind of goal-oriented binarization. 
Instead of detecting all foreground elements, it restricts itself to those elements which are of interest for the specific task.
The second stage performs a superpixel (SP) extraction on the first stage's output. These SPs are further clustered to build baselines. In the following, the problem of baseline detection is formulated. 
Afterwards, a detailed description of the proposed ARU-Net is given. Finally, the SP extraction and clustering approach is described.
\subsection{Problem Statement}
\label{ssec:ps}
We will introduce the problem of baseline detection in a formal way by defining all necessary termini and notation.
Within this work we follow the definition of a baseline given in \cite{diem_markus_2017_257972}: 
\begin{defi}[baseline]
\label{def:bl}
 A \textit{baseline} is defined in the typographical sense as the
virtual line where most characters rest upon and descenders extend below.
\end{defi}
Hence, each baseline can be represented by a polygonal chain. The infinite set $\mathcal{P}$ of all possible polygonal chains is called \textit{polygonal chain space}. Within this work we limit ourselves towards gray-scale images. The set of all possible (gray-scale) images $\mathcal{I}=\bigcup_{h,w\in\mathbb{N}} \left[0,1\right]^{h\times w}$ is called \textit{image space}. If the colored image is available, we usually use this one for visualization even though it is converted to its gray-scale version for calculations. For visualization purposes a pixel intensity value of $1$ means white and $0$ means black. $I_h$ denotes the height of image $I$, $I_w$ denotes the width, analogously. 
\begin{defi}[baseline detector, baseline hypothesis]
 We call a function $b:\mathcal{I}\rightarrow\mathfrak{P}(\mathcal{P})$ which maps each image to a subset of $\mathcal{P}$ a \textit{baseline detector}. The set of all baseline detectors is denoted by $\mathcal{B}$. 
 The output of $b$ for a certain image $I$ is called \textit{baseline hypothesis}.
\end{defi}
\begin{defi}[baseline ground truth]
 The set  $\mathcal{G}_I\subset\mathcal{P}$ of polygonal chains representing the baselines of an image $I$ (possibly annotated by a human operator) is called \textit{baseline ground truth} (for image $I$).
\end{defi}
 Def.~\ref{def:bl} allows for some baseline variety. Hence, there is not the one unique and correct ground truth for an image. 
 Therefore, ground truth information is always biased by its creator. This has to be taken into account for the evaluation process as well as for the baseline detector design.
\begin{defi}[similarity score]
 A function $\langle\cdot ,\cdot  \rangle_{\mu}:\mathcal{P}\times\mathcal{P}\rightarrow \left[0,1\right]$ assigning a scalar value to each pair of baseline ground truth and baseline hypothesis polygonal chain sets is called \textit{similarity score}.
\end{defi}
 A value of $1.0$ indicates that two polygonal chains are regarded as equal.
 Within this work we follow the similarity score introduced in \cite{gruning2017read}: We measure the accuracy of a baseline detector in terms of the F-value, see \cite{gruning2017read} for a detailed introduction. 

The problem tackled in this work can now be formulated as follows:
Suppose there are two sets of images along with their baseline ground truth information
$\mathcal{T}_{train}=\{(I,\mathcal{G}_I)_i\ |\ i=1,...,n\}$ and  $\mathcal{T}_{test}=\{(I,\mathcal{G}_I)_i\ |\ i=1,...,m\}$.
We aim for a design of a baseline detector $b^*$ given $\mathcal{T}_{train}$ which solves
\begin{align*}
  b^*=\argmax_{b\in\mathcal{B}}\sum_{(I,\mathcal{G}_I)\in\mathcal{T}_{test}} \langle \mathcal{G}_I , b(I)  \rangle_{\mu}.
\end{align*}
In the design phase of $b^*$ the set $\mathcal{T}_{test}$ is unknown and one is allowed to use solely $\mathcal{T}_{train}$. 
Hence, one has to ensure that $b^*$ generalizes well from $\mathcal{T}_{train}$ to $\mathcal{T}_{test}$.

Since the proposed design consists of two stages and the first stage relies on deep learning techniques, 
an adaptation to a differently biased ground truth (produced by a different annotator) can be done easily by retraining the first stage without any fine tuning done by experts. 
\subsection{Stage I: ARU-Net}
\label{ssec:aru}
Typically, layout analysis algorithms directly work on the input image $I$ or on a binarized version of it \cite{Arvanitopoulos2014,Ryu2014,Nicolaou2009,Saabni2014,Garz2012,gr2017}.
Instead, we employ a more goal-oriented transformation of the input image utilizing a neural network, which is trained in a supervised manner to assign a certain class to each pixel like in \cite{Ronneberger2015,Long2015,Noh2015}. 
This is often referred to as pixel labeling or semantic segmentation. 
We will introduce the problem of pixel labeling utilizing hierarchical neural networks, followed by a description of the proposed ARU-Net architecture.
\subsubsection{Pixel Labeling -- Problem Formulation}
\begin{defi}[neural pixel labeler]
\label{def:pl}
 A \textit{neural pixel labeler} (NPL) for the classes $\mathcal{C}=\left\{c_1,...,c_n\right\}$ is a hierarchical neural network $\Phi(\ \cdot\ ;\boldsymbol{w}):\mathcal{I} \rightarrow \mathcal{I}^{\left|\mathcal{C}\right|}$. The NPL is  parametrized by $\boldsymbol{w}\in\mathbb{R}^N$. 
 For $I\in\mathcal{I}$ it performs a prediction over all pixels and all possible classes
 $\Phi(I;\boldsymbol{w})=C\in\left[0,1\right]^{I_h\times I_w \times \left|\mathcal{C}\right|}$,
 where $C$ sums to one over all classes and for all coordinates.
\end{defi}
 $C(:,:,c)=C_{:,:,c}\in\mathcal{I}$ denotes the image which encodes the pixel-wise prediction (confidence) for the $c-$th class.
\begin{defi}[pixel ground truth]
 A cartesian product $G_I\in\mathcal{I}^{\left|\mathcal{C}\right|}$ is called \textit{pixel ground truth} (for image $I$) if it assigns exactly one class (one-hot-encoding) to each pixel.
\end{defi}
Following the problem formulation of Section~\ref{ssec:ps}, we aim for an NPL, which was tuned on a training set and optimally performs on a test set. 
Assume there are training and test sets as stated above, but with pixel ground truth information instead of baseline ground truth information, which are denoted by $\widetilde{\mathcal{T}}_{train}$ and $\widetilde{\mathcal{T}}_{test}$.
The performance of an NPL is evaluated in terms of the cross-entropy between the predicted and the ground truth distribution. The cross-entropy can also be motivated by a maximum likelihood estimation. This results in the cross-entropy loss function.
\begin{defi}[loss function]
 Let $\widetilde{\mathcal{T}}$ be a set of images along with their pixel ground truth and $\Phi(\ \cdot\ ;\boldsymbol{w})$ is an NPL. The performance of $\Phi$ 
 on $\widetilde{\mathcal{T}}$ is evaluated in terms of the (cross-entropy) \textit{loss function}
 \begin{align*}
  -\sum_{(I,G_I)\in\widetilde{\mathcal{T}}}\sum_{y=1}^{I_h}\sum_{x=1}^{I_w}\sum_{c=1}^{\left|\mathcal{C}\right|}G_I(y,x,c)\ln{\Phi(I;\boldsymbol{w})_{y,x,c}}.
 \end{align*}
\end{defi}
To improve the performance of the NPL on the training set, one can calculate the loss function's gradient with respect to the model parameters using the well-known technique of backpropagation \cite{Rumelhart1986}. 
The gradient is used to update the model parameters by gradient descent:
$\boldsymbol{w} \gets \boldsymbol{w} - \tau\cdot\frac{\partial L}{\partial \boldsymbol{w}}(\Phi,\widetilde{\mathcal{T}}_{train})$
with a \textit{learning rate} $\tau$. This is repeated to successively adapt the NPL. The process of adapting the model by minimizing its loss is called \textit{training}.
Since one does not aim for a minimization of the loss on the training set, the system has to generalize to achieve high quality results on the test set as well.
To stabilize training, avoid over-fitting, improve generalization, ... dozens of techniques to improve the simple update rule which is stated above were introduced within the last years. 
Since the introduction of these is beyond the scope of this work, we refer to \cite{Goodfellow-et-al-2016}. Details on techniques used within this work are given in Sec.~\ref{sec:exp}.
\subsubsection{ARU-Net -- Architecture}
The ARU-Net is a special form of an NPL and is described in this section. We omit a formal introduction of the used neural network components and concepts and refer to the above mentioned literature.
Within the last few years, different architectures were proposed for the pixel labeling task. 
Most of them are based on Convolutional Neural Networks (CNNs) \cite{LeCun1998}. A direct application of CNNs for semantic segmentation
is presented in \cite{Long2015}. The presented Fully Convolutional Network (FCN) combines local features to produce more meaningful high level features using pooling layers.
Pooling reduces the spatial dimension. Thus, the result suffers from a coarse resolution. Noh et al. \cite{Noh2015} tackle this problem by applying a deconvolutional network on the subsampled output of the FCN. The U-Net 
proposed in \cite{Ronneberger2015} furthermore introduces shortcuts between layers of the same spatial dimension. This allows for an easier combination of local low level features and global higher-level features. Additionally, error propagation for deep structures is facilitated and the so-called vanishing gradient problems \cite{pmlr-v9-glorot10a} are reduced.
The U-Net is the basis for the proposed ARU-Net. We extend the U-Net by two more key concepts -- spatial attention (A) and depth (residual structure (R)) to be described below. Remarkably, in contrast to the U-Net proposed in \cite{Ronneberger2015}, we perform border padding. 
Hence, the spatial dimensions in each scale space of the U-Net are all the same, see Fig.~\ref{fig:unet} for a schematic representation of an U-Net. The output of the U-Net thus is a feature map ($Z$ features in Fig.~\ref{fig:unet}) of the same spatial dimension as the input. 
 Hence, the U-Net becomes an NPL as defined in Def.~\ref{def:pl} by adding a convolutional (to get pixel-wise predictions) softmax classifier on top which distinguishes between the different classes of $\mathcal{C}$.
 \tikzstyle{conv} = [color=blue!60,very thick,-stealth]
\tikzstyle{pool} = [color=black!60,thick,-angle 90]
\tikzstyle{up} = [color=black!60,thick,-latex]
\tikzstyle{copy} = [color=black!40,thick,-angle 90]
\newcommand{\drawUNet}[4]
{
    \begin{tikzpicture}
		\node[color=red] at (0,0.5*#1pt+3) {\tiny $1$};
		\node[color=red] at (0.5*#2+#3pt,0.5*#1pt+3) {\tiny $Z$};
		\node[color=red] at (1.5*#2+2*#3pt,0.5*#1pt+3) {\tiny $Z$};
		\node[color=red] at (3.0*#2+3*#3pt,-0.5*#1-#4pt+3) {\tiny $2Z$};
		\node[color=red] at (5.0*#2+4*#3pt,-0.5*#1-#4pt+3) {\tiny $2Z$};
		\node[color=red] at (8.0*#2+5*#3pt,-1.0*#1-2*#4pt+3) {\tiny $4Z$};
		\node[color=red] at (12.0*#2+6*#3pt,-1.0*#1-2*#4pt+3) {\tiny $4Z$};		
		\node[color=red] at (18.0*#2+7*#3pt,-1.25*#1-3*#4pt+3) {\tiny $8Z$};
		\node[color=red] at (24.0*#2+8*#3pt,-1.0*#1-2*#4pt+3) {\tiny $8Z$};
		\node[color=red] at (30.0*#2+9*#3pt,-1.0*#1-2*#4pt+3) {\tiny $4Z$};		
		\node[color=red] at (33.0*#2+10*#3pt,-0.5*#1-#4pt+3) {\tiny $4Z$};
		\node[color=red] at (36.0*#2+11*#3pt,-0.5*#1-#4pt+3) {\tiny $2Z$};		
		\node[color=red] at (37.5*#2+12*#3pt,0.5*#1pt+3) {\tiny $2Z$};
		\node[color=red] at (39.0*#2+13*#3pt,0.5*#1pt+3) {\tiny $Z$};
		\node[color=red] at (40.0*#2+14*#3pt,0.5*#1pt+3) {\tiny $Z$};		
		
		\node[color=black,rotate=90] at (-4pt,0) {\tiny Input};
		\node[color=black,rotate=90] at (40.5*#2+14*#3pt+4pt,0) {\tiny Output};

		\node[color=black] at (-#4pt,0) {\large I};
		\node[color=black] at (-#4pt,-0.75*#1-#4pt) {\large II};
		\node[color=black] at (-#4pt,-1.125*#1-2.0*#4pt) {\large III};
		\node[color=black] at (-#4pt,-1.3125*#1-3.0*#4pt) {\large IV};
    
		\draw [dashed,color=black] (-#4pt,-0.5*#1-0.5*#4pt) -- (45*#2+15*#3pt,-0.5*#1-0.5*#4pt);
		\draw [dashed,color=black] (-#4pt,-#1-1.5*#4pt) -- (45*#2+15*#3pt,-#1-1.5*#4pt);
		\draw [dashed,color=black] (-#4pt,-1.25*#1-2.5*#4pt) -- (45*#2+15*#3pt,-1.25*#1-2.5*#4pt);
    
		\node (CNN1) at (0.5*#2+#3pt,0) [color=green!60,rounded corners,draw,rectangle,inner sep=0pt,minimum width=3.5*#2+3*#3,minimum height=#1+14pt] {};
		\node (CNN2) at (3.5*#2+3*#3pt,-0.75*#1-#4pt) [color=green!60,rounded corners,draw,rectangle,inner sep=0pt,minimum width=6.0*#2+3*#3,minimum height=0.5*#1+14pt] {};
		\node (CNN3) at (9.0*#2+5*#3pt,-1.125*#1-2.0*#4pt) [color=green!60,rounded corners,draw,rectangle,inner sep=0pt,minimum width=11.0*#2+3*#3,minimum height=0.25*#1+14pt] {};
		\node (CNN4) at (20.0*#2+7*#3pt,-1.3125*#1-3.0*#4pt) [color=green!60,rounded corners,draw,rectangle,inner sep=0pt,minimum width=21.0*#2+3*#3,minimum height=0.125*#1+14pt] {};
		\node (CNN5) at (28.0*#2+9*#3pt,-1.125*#1-2.0*#4pt) [color=green!60,rounded corners,draw,rectangle,inner sep=0pt,minimum width=17.0*#2+3*#3,minimum height=0.25*#1+14pt] {};
		\node (CNN6) at (35.0*#2+11*#3pt,-0.75*#1-#4pt) [color=green!60,rounded corners,draw,rectangle,inner sep=0pt,minimum width=9.0*#2+3*#3,minimum height=0.5*#1+14pt] {};
		\node (CNN7) at (38.5*#2+13*#3pt,0) [color=green!60,rounded corners,draw,rectangle,inner sep=0pt,minimum width=5.0*#2+3*#3+4pt,minimum height=#1+14pt] {};

		\node (I) at (0,0) [draw,rectangle,inner sep=0pt,minimum width=0.0cm,minimum height=#1pt] {};
		\node (d1a) at (0.5*#2+#3pt,0) [draw,rectangle,inner sep=0pt,minimum width=#2,minimum height=#1] {};
		\node (d1b) at (1.5*#2+2*#3pt,0) [draw,pattern color=black!20,rectangle,inner sep=0pt,minimum width=#2,minimum height=#1] {};
		
		\node (d2a) at (1.5*#2+2*#3pt,-0.75*#1-#4pt) [draw,rectangle,inner sep=0pt,minimum width=#2,minimum height=0.5*#1] {};
		\node (d2b) at (3.0*#2+3*#3pt,-0.75*#1-#4pt) [draw,rectangle,inner sep=0pt,minimum width=2.0*#2,minimum height=0.5*#1] {};
		\node (d2c) at (5.0*#2+4*#3pt,-0.75*#1-#4pt) [draw,pattern color=black!20,rectangle,inner sep=0pt,minimum width=2.0*#2,minimum height=0.5*#1] {};
		
		\node (d3a) at (5.0*#2+4*#3pt,-1.125*#1-2.0*#4pt) [draw,rectangle,inner sep=0pt,minimum width=2.0*#2,minimum height=0.25*#1] {};
		\node (d3b) at (8.0*#2+5*#3pt,-1.125*#1-2.0*#4pt) [draw,rectangle,inner sep=0pt,minimum width=4.0*#2,minimum height=0.25*#1] {};
		\node (d3c) at (12.0*#2+6*#3pt,-1.125*#1-2.0*#4pt) [draw,pattern color=black!20,rectangle,inner sep=0pt,minimum width=4.0*#2,minimum height=0.25*#1] {};
		
		\node (d4a) at (12.0*#2+6*#3pt,-1.3125*#1-3.0*#4pt) [draw,rectangle,inner sep=0pt,minimum width=4.0*#2,minimum height=0.125*#1] {};
		\node (d4b) at (18.0*#2+7*#3pt,-1.3125*#1-3.0*#4pt) [draw,rectangle,inner sep=0pt,minimum width=8.0*#2,minimum height=0.125*#1] {};
		\node (d4c) at (26.0*#2+8*#3pt,-1.3125*#1-3.0*#4pt) [draw,rectangle,inner sep=0pt,minimum width=8.0*#2,minimum height=0.125*#1] {};	
		
		\node (u3a1) at (26.0*#2+8*#3pt,-1.125*#1-2.0*#4pt) [rectangle,inner sep=0pt,minimum width=4.0*#2,minimum height=0.25*#1] {};
		\node (u3a2) at (22.0*#2+8*#3pt,-1.125*#1-2.0*#4pt) [pattern=north east lines,pattern color=black!40,rectangle,inner sep=0pt,minimum width=4.0*#2,minimum height=0.25*#1] {};
		\node (u3a) at (24.0*#2+8*#3pt,-1.125*#1-2.0*#4pt) [draw,rectangle,inner sep=0pt,minimum width=8.0*#2,minimum height=0.25*#1] {};
		\node (u3b) at (30.0*#2+9*#3pt,-1.125*#1-2.0*#4pt) [draw,rectangle,inner sep=0pt,minimum width=4.0*#2,minimum height=0.25*#1] {};
		\node (u3c) at (34.0*#2+10*#3pt,-1.125*#1-2.0*#4pt) [draw,rectangle,inner sep=0pt,minimum width=4.0*#2,minimum height=0.25*#1] {};

		\node (u2a1) at (34.0*#2+10*#3pt,-0.75*#1-#4pt) [rectangle,inner sep=0pt,minimum width=2.0*#2,minimum height=0.5*#1] {};
		\node (u2a2) at (32.0*#2+10*#3pt,-0.75*#1-#4pt) [pattern=north east lines,pattern color=black!40,rectangle,inner sep=0pt,minimum width=2.0*#2,minimum height=0.5*#1] {};
		\node (u2a) at (33.0*#2+10*#3pt,-0.75*#1-#4pt) [draw,rectangle,inner sep=0pt,minimum width=4.0*#2,minimum height=0.5*#1] {};
		\node (u2b) at (36.0*#2+11*#3pt,-0.75*#1-#4pt) [draw,rectangle,inner sep=0pt,minimum width=2.0*#2,minimum height=0.5*#1] {};
		\node (u2c) at (38.0*#2+12*#3pt,-0.75*#1-#4pt) [draw,rectangle,inner sep=0pt,minimum width=2.0*#2,minimum height=0.5*#1] {};

		\node (u1a1) at (38.0*#2+12*#3pt,0) [rectangle,inner sep=0pt,minimum width=#2,minimum height=#1] {};
		\node (u1a2) at (37.0*#2+12*#3pt,0) [pattern=north east lines,pattern color=black!40,rectangle,inner sep=0pt,minimum width=#2,minimum height=#1] {};
		\node (u1a) at (37.5*#2+12*#3pt,0) [draw,rectangle,inner sep=0pt,minimum width=2.0*#2,minimum height=#1] {};
		\node (u1b) at (39.0*#2+13*#3pt,0) [draw,rectangle,inner sep=0pt,minimum width=#2,minimum height=#1] {};
		\node (u1c) at (40.0*#2+14*#3pt,0) [draw,rectangle,inner sep=0pt,minimum width=#2,minimum height=#1] {};

		\draw [conv] (I)--(d1a);
		\draw [conv] (d1a)--(d1b);
		\draw [conv] (d2a)--(d2b);
		\draw [conv] (d2b)--(d2c);
		\draw [conv] (d3a)--(d3b);
		\draw [conv] (d3b)--(d3c);
		\draw [conv] (d4a)--(d4b);
		\draw [conv] (d4b)--(d4c);
		\draw [conv] (u3a)--(u3b);
		\draw [conv] (u3b)--(u3c);
		\draw [conv] (u2a)--(u2b);
		\draw [conv] (u2b)--(u2c);
		\draw [conv] (u1a)--(u1b);
		\draw [conv] (u1b)--(u1c);
		
		
		\draw [pool] (d1b)--(d2a);
		\draw [pool] (d2c)--(d3a);
		\draw [pool] (d3c)--(d4a);
		
		\draw [up] (d4c)--(u3a1);
		\draw [up] (u3c)--(u2a1);
		\draw [up] (u2c)--(u1a1);
		
		\draw [copy] (d3c)--(u3a2);
		\draw [copy] (d2c)--(u2a2);
		\draw [copy] (d1b)--(u1a2);
		
		\node[color=black!60] at (17.0*#2+7*#3pt,-1.125*#1-2.0*#4+4pt) {\tiny Ident};
		\node[color=black!60] at (17.0*#2+7*#3pt,-0.75*#1-#4+4pt) {\tiny Ident};
		\node[color=black!60] at (17.0*#2+7*#3pt,4pt) {\tiny Ident};

		\draw [conv] (-#4-5pt,-1.3125*#1-4.5*#4pt+8pt)--(-#4+5pt,-1.3125*#1-4.5*#4pt+8pt);
		\node[] at (-#4+35pt,-1.3125*#1-4.5*#4pt+8pt) {\small Conv+Act};
		
		\draw [pool] (-#4+70pt,-1.3125*#1-4.5*#4+7pt+8pt)--(-#4+70pt,-1.3125*#1-4.5*#4-7pt+8pt);
		\node[] at (-#4+97pt,-1.3125*#1-4.5*#4pt+8pt) {\small Max Pool};

		\draw [up] (-#4+127pt,-1.3125*#1-4.5*#4-7pt+8pt)--(-#4+127pt,-1.3125*#1-4.5*#4+7pt+8pt);
		\node[] at (-#4+160pt,-1.3125*#1-4.5*#4pt+8pt) {\small Deconv+Act};
		
		
		\node () at (-#4-1pt,-1.3125*#1-4.5*#4pt-20pt+8pt) [color=green!60,rounded corners,draw,rectangle,inner sep=0pt,minimum width=15pt,minimum height=15pt] {};
		\node () at (-#4+55pt,-1.3125*#1-4.5*#4pt-20+8pt) [] {2 Layer CNN Block};
				
	\end{tikzpicture}
}
\begin{figure}[ht]
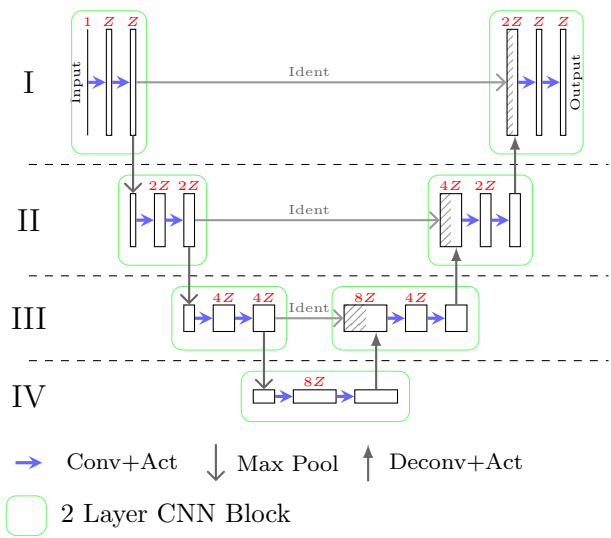

	\centering
	\drawUNet{40}{2}{7}{22}
	\caption{\textbf{U-Net --} The input is an image of arbitrary spatial dimension. "Act" is the activation function thus the rectangles represent sets of activation maps. Each rectangle represents a $3$-dim array ($\in\mathbb{R}^{h\times w\times z}$). Within each scale space (roman numbers) the feature map widths and heights are constant (encoded by the height of the rectangles). The number of feature maps $Z$ is pictured by the width of the rectangles. 
	Between adjacent scale spaces the spatial dimension decreases by a certain factor ($2$ in the figure) and the representative depth (number of feature maps) increases by the same factor.}
	\label{fig:unet}
\end{figure}
  \begin{rem} 
  	\label{rem:npl}
  	If the presented architectures are used for the pixel labeling task, it is implicitly assumed that such a classifier is always added to generate per class confidences at pixel level.
  \end{rem}
He et al. \cite{He2016} introduce very deep neural networks which are still trainable and yield state-of-the-art results. This is achieved using so-called residual blocks. Residual blocks introduce shortcuts, which enable the error backpropagation and identity propagation even for very deep structures. Hence, the vanishing gradient problems are reduced \cite{He2016}. There are various different forms of residual blocks.
The one used within this work is depicted in Fig.~\ref{fig:res}. 
\begin{defi}[RU-Net]
 An \textit{RU-Net} is an U-Net with residual blocks.
\end{defi}
That means, each of the $2$ layer CNN blocks in Fig.~\ref{fig:unet} is replaced by a residual block as in Fig.~\ref{fig:res}.
\tikzstyle{convA} = [color=blue!60,very thick,-stealth]
\tikzstyle{conv} = [color=red!60,very thick,-stealth]
\tikzstyle{act} = [color=black,thick,-latex]
\tikzstyle{copy} = [color=black!40,thick,->]
\newcommand{\drawResBlock}[4]
{
    \begin{tikzpicture}
    
		\node (I) at (0,0) [draw,rectangle,inner sep=0pt,minimum width=0,minimum height=#1] {};
		\node (I2) at (2*#2pt,0) [draw,rectangle,inner sep=0pt,minimum width=0,minimum height=#1] {};
		\draw [densely dotted,color=black] (0,-0.5*#1pt) -- (2*#2pt,-0.5*#1pt);
		\draw [densely dotted,color=black] (0,0.5*#1pt) -- (2*#2pt,0.5*#1pt);
		\node[color=black,rotate=90] at (-4pt,0) {\tiny Input};
		\node (d1a) at (2.5*#2+#3pt,0) [draw,rectangle,dashed,inner sep=0pt,minimum width=#2,minimum height=#1] {};
		\node[color=red] at (2.5*#2+#3pt,0.5*#1pt+3) {\tiny Z};
		\draw [conv] (I2)--(d1a);	
		\node (c1) at (3.0*#2+1.5*#3+0.5*#4pt,0) [draw,circle,inner sep=0pt,minimum width=#4,minimum height=#4] {};
		\node (c1b) at (3.0*#2+1.5*#3+0.5*#4pt,0.5*#1+8pt) [inner sep=0pt,minimum width=0,minimum height=0] {};				
		\draw [copy] (d1a)--(c1);		
		\node (d2a) at (3.5*#2+2.0*#3+#4pt,0) [draw,rectangle,inner sep=0pt,minimum width=#2,minimum height=#1] {};
		\node[color=red] at (3.5*#2+2.0*#3+#4pt,0.5*#1pt+3) {\tiny Z};
		\draw [act] (c1)--(d2a);		
		\node (h1) at (4.0*#2+3.0*#3+#4pt,0) [inner sep=0pt,minimum width=0,minimum height=0] {};			
		\node (h2) at (4.0*#2+4.0*#3+#4pt,0) [inner sep=0pt,minimum width=0,minimum height=0] {};
		\node (h3) at (4.0*#2+3.0*#3+2.0*#4pt,0) [inner sep=0pt,minimum width=0,minimum height=0] {};
		\node (h4) at (4.0*#2+4.0*#3pt,0) [inner sep=0pt,minimum width=0,minimum height=0] {};
		\draw [dotted,very thick,-] (h3)--(h4);
		\draw [convA] (d2a)--(h1);				
		\node (d4a) at (4.5*#2+5.0*#3+#4pt,0) [draw,rectangle,inner sep=0pt,minimum width=#2,minimum height=#1] {};
		\node[color=red] at (4.5*#2+5.0*#3+#4pt,0.5*#1pt+3) {\tiny Z};
		\draw [convA] (h2)--(d4a);				
		\node (d5a) at (5.5*#2+6.0*#3+#4pt,0) [draw,dashed,rectangle,inner sep=0pt,minimum width=#2,minimum height=#1] {};
		\node[color=red] at (5.5*#2+6.0*#3+#4pt,0.5*#1pt+3) {\tiny Z};
		\draw [conv] (d4a)--(d5a);  
		\node (c2) at (6.0*#2+6.5*#3+2.0*#4pt,0) [draw,circle,inner sep=0pt,minimum width=0,minimum height=0] {\tiny $+$};
		\node (c2b) at (6.0*#2+6.5*#3+2.0*#4pt,0.5*#1+8pt) [inner sep=0pt,minimum width=0,minimum height=0] {};
		\draw [copy] (c1)|-(c1b)--(c2b)-|(c2);
		\draw [copy] (d5a)--(c2);				
		\node (d6a) at (6.5*#2+7.0*#3+3.0*#4pt,0) [draw,rectangle,inner sep=0pt,minimum width=#2,minimum height=#1] {};
		\node[color=red] at (6.5*#2+7.0*#3+3.0*#4pt,0.5*#1pt+3) {\tiny Z};
		\draw[act] (c2)--(d6a);		
		\node[color=black,rotate=90] at (7.0*#2+7.0*#3+3.0*#4+4pt,0) {\tiny Output};				
		\node (CNN1) at (3.25*#2+3.5*#3+1.5*#4pt,0) [color=green!60,rounded corners,draw,rectangle,inner sep=0pt,minimum width=6.5*#2+7.0*#3+3.0*#4+30pt,minimum height=#1+30pt] {};		
		\draw [convA] (-15pt,-0.5*#1-30pt)--(-5pt,-0.5*#1-30pt);
		\node[] at (+19pt,-0.5*#1-30pt) {\small Conv+Act};		
		\draw [conv] (50pt,-0.5*#1-30pt)--(60pt,-0.5*#1-30pt);
		\node[] at (73pt,-0.5*#1-30pt) {\small Conv};		
		\draw [act] (93pt,-0.5*#1-30pt)--(103pt,-0.5*#1-30pt);
		\node[] at (111pt,-0.5*#1-30pt) {\small Act};		
		\draw [copy] (127pt,-0.5*#1-30pt)--(137pt,-0.5*#1-30pt);
		\node[] at (155pt,-0.5*#1-30pt) {\small Identity};				
		\node (d6a) at (185pt,-0.5*#1-30pt) [draw,rectangle,dashed,inner sep=0pt,minimum width=8pt,minimum height=12pt] {};
		\node[] at (206pt,-0.5*#1-30pt) {\small Logits};

	\end{tikzpicture}
}
\begin{figure}[ht]
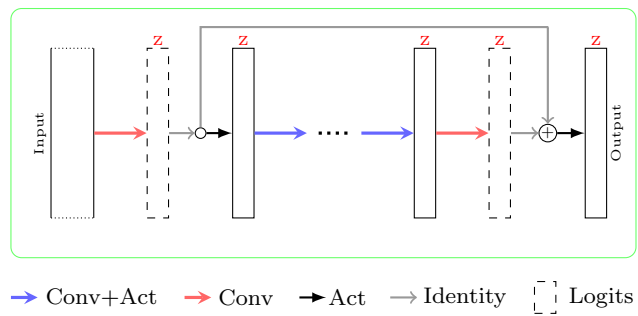

	\centering
	\drawResBlock{64}{8}{20}{4}
	\caption{\textbf{Residual Block --} The input is convolved and the resulting $3$-dim array (the maps before passed through an acitvation function are referred to as logits) is used twice. At the first branch it is passed through the activation function and further processed by several convolution layers.
	At the second branch it is directly fed into a summation node. After a point-wise summation of the two logit maps an activation function is applied. The shortcut enables for an easy identity propagation and error backpropagation. Arbitrarily many inner layers are possible.}
	\label{fig:res}
\end{figure}

To explicitly incorporate the potential to handle various font sizes, especially mixed font sizes on a single page, we introduce a pixel-wise (spatial) attention mechanism. 
For this purpose, we introduce an attention network (A-Net). The A-Net is a multi-layer CNN which generates a single output feature map.
The A-Net will be applied along with the RU-Net at different scales, the same network weights are used on all scales (weight sharing). Specially, a scale pyramid is built by downscaling the input image $I = I_1$ several times. The resulting (scaled) images $I_1,\ I_2,\ I_4,\ I_8,...,\ I_s$ (subscripts denote the scaling factors) are fed into the RU-Net and the A-Net.  
Trainable deconvolutional layers (of corresponding scales) are applied on the outputs of the RU- and the A-Net to obtain feature maps of spatial dimensions equal to the inputs. $A_1,...,A_s$ denote the up-sampled feature 
maps of the A-Net, $RU_1,...,RU_s$ of the RU-Net, respectively.
After applying a pixel-wise softmax normalization for the attention maps
\begin{align*}
 \widehat{A}_i(y,x)=\frac{\exp(A_i(y,x))}{\sum_{j\in\{1,2,...,s\}} \exp(A_j(y,x))}
\end{align*}
the normalized attention maps $\widehat{A}_i$ sum to one (pixel-wise). 
The feature maps $RU_i$ are combined following
\begin{align*}
 ARU=\sum_{i\in\{1,2,...,s\}} RU_i\odot \widehat{A}_i,
\end{align*}
where $\odot$ is the Hadamard product. $ARU$ is the input for the classifier to build a NPL, see Rem.~\ref{rem:npl}.
\tikzstyle{down} = [color=black!60,thick,-angle 90]
\tikzstyle{copy} = [color=black!80,thick,->]
\tikzstyle{ru} = [color=red!60,thick,->]
\tikzstyle{att} = [color=blue!60,thick,->]
\newlength{\myheight}
\setlength{\myheight}{56pt}
\newcommand{\drawARU}
{
    \begin{tikzpicture}
		\node (pic1) at (0,0) {\includegraphics[height=\myheight]{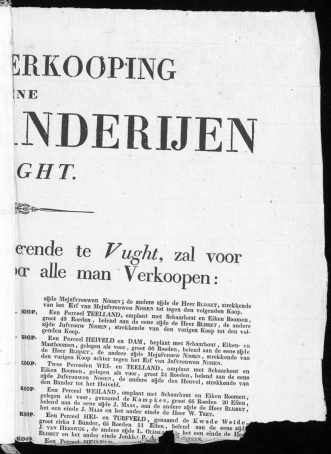}};
		
		\node (pic2) at (0,-1.1\myheight) {\includegraphics[height=0.5\myheight]{img2.png}};
		
		\node (pic4) at (0,-2.2\myheight) {\includegraphics[height=0.25\myheight]{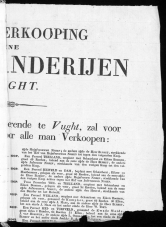}};
		
		\draw [down] (pic1)--(pic2);
		\draw [down] (pic2)--(pic4);
		
		\node (att1) at (3\myheight,0) {\includegraphics[height=\myheight]{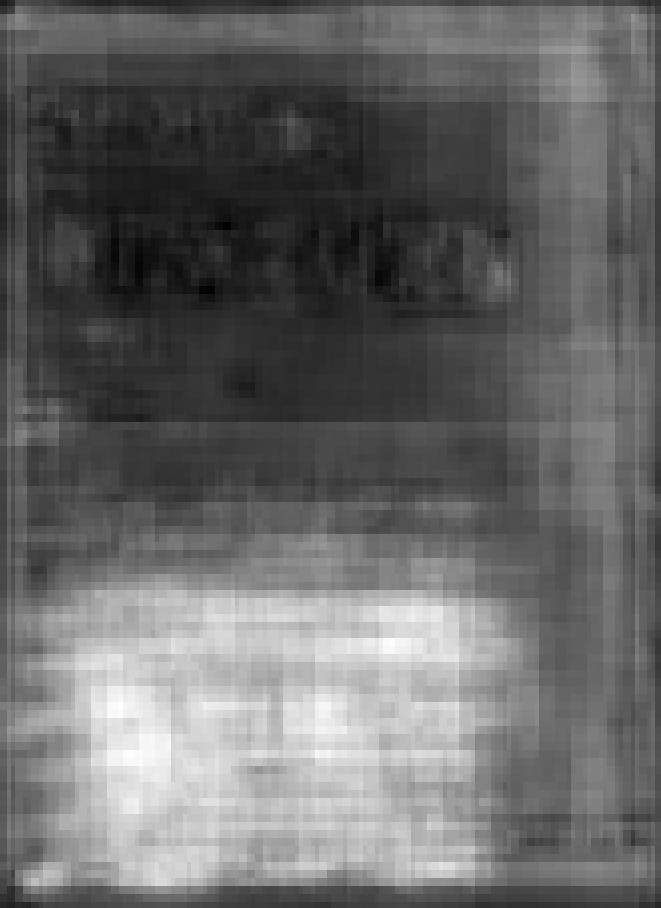}};
		
		\node (m1) at (3.6\myheight,0) [inner sep=0,outer sep=0,minimum size=0]{$\odot$};
		
		\node (att2) at (3\myheight,-1.1\myheight) {\includegraphics[height=\myheight]{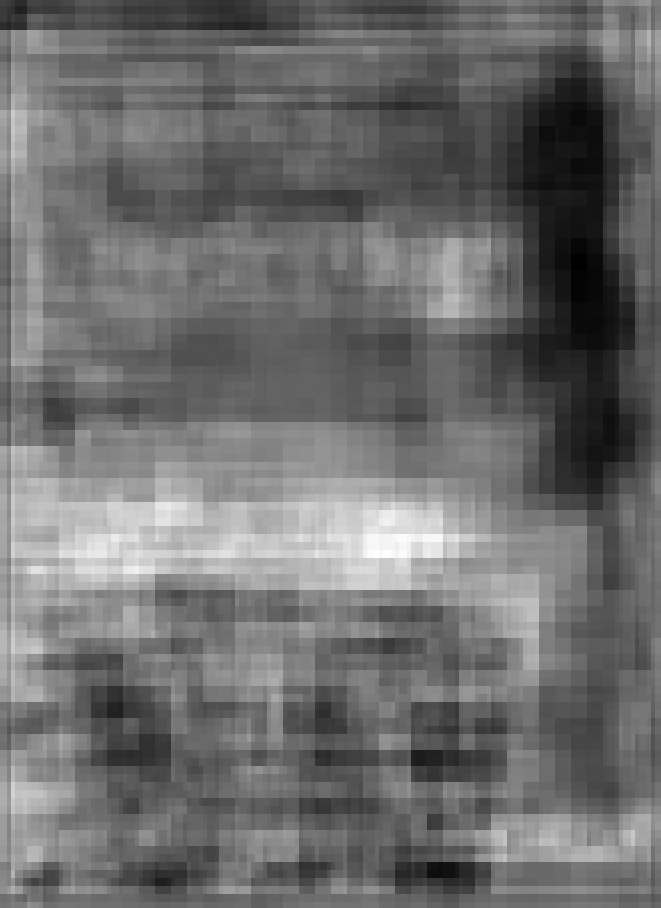}};
		
		\node (m2) at (3.6\myheight,-1.1\myheight) [inner sep=0,outer sep=0,minimum size=0]{$\odot$};
		
		\node (att4) at (3\myheight,-2.2\myheight) {\includegraphics[height=\myheight]{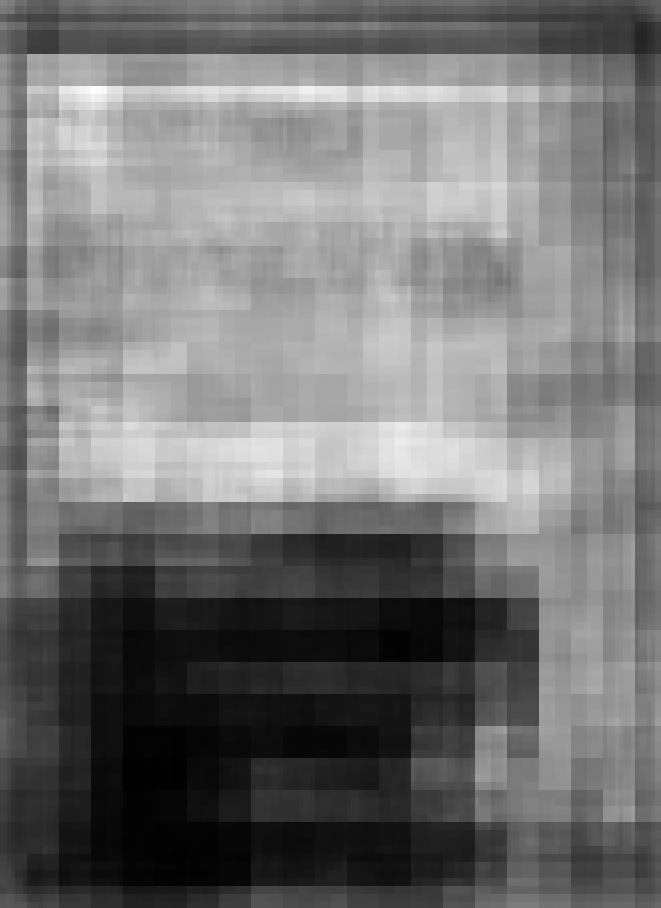}};
		
		\node (p4) at (3.9\myheight,-1.1\myheight) [draw,circle,inner sep=0,minimum size=6pt]{\tiny $+$};
		
		\node (C1) at (4.25\myheight,-1.1\myheight) [rotate=90,draw,rectangle,rounded corners,minimum width=1.6\myheight] {\tiny Classification};
		
		\node (m4) at (3.6\myheight,-2.2\myheight)[inner sep=0,outer sep=0,minimum size=0]{$\odot$};
		
		\node (res1) at (5\myheight,-0.55\myheight) {\includegraphics[height=\myheight]{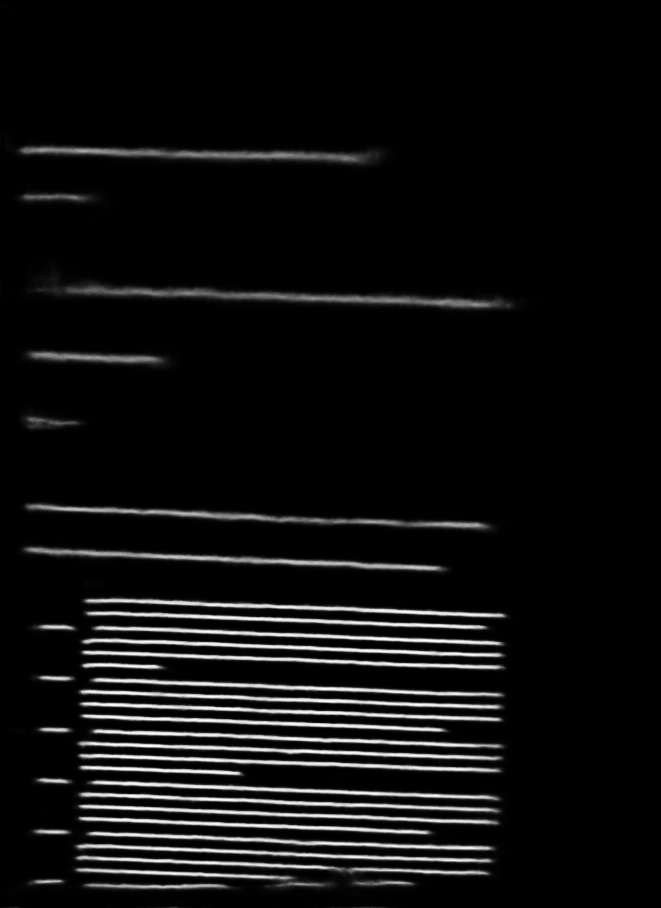}};
		\node (res2) at (5\myheight,-1.65\myheight) {\includegraphics[height=\myheight]{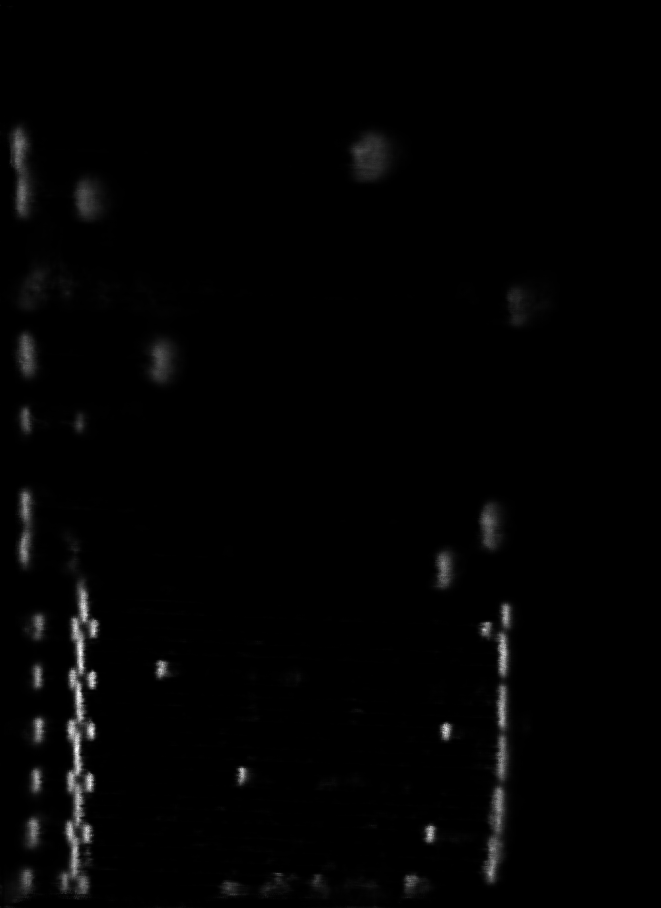}};
		
		\draw [copy] (m1)--(3.9\myheight,0)--(p4);
		\draw [copy] (m4)--(3.9\myheight,-2.2\myheight)--(p4);
		
		\draw [copy] (C1.south|-res1.west)--(res1);
		\draw [copy] (C1.south|-res2.west)--(res2);
		
		\draw [copy] (m2)--(p4);
		\draw [copy] (p4)--(C1);
		
		\node (A1) at (\myheight,0.25\myheight) [draw,rectangle,rounded corners,minimum width=50pt] {\tiny  A-Net};
		\node (R1) at (\myheight,-0.25\myheight) [draw,rectangle,rounded corners,minimum width=50pt] {\tiny  RU-Net};
		
		\node (A2) at (\myheight,-0.85\myheight) [draw,rectangle,rounded corners,minimum width=50pt] {\tiny  A-Net};
		\node (R2) at (\myheight,-1.35\myheight) [draw,rectangle,rounded corners,minimum width=50pt] {\tiny  RU-Net};
		
		\node (A4) at (\myheight,-1.95\myheight) [draw,rectangle,rounded corners,minimum width=50pt] {\tiny  A-Net};
		\node (R4) at (\myheight,-2.45\myheight) [draw,rectangle,rounded corners,minimum width=50pt] {\tiny  RU-Net};
		
		\draw [copy] (pic1.0)--(0.45\myheight,0)--(0.45\myheight,0.25\myheight)--(A1.180);
		\draw [copy] (pic1.0)--(0.45\myheight,0)--(0.45\myheight,-0.25\myheight)--(R1.180);
		
		\draw [copy] (pic2.0)--(0.45\myheight,-1.1\myheight)--(0.45\myheight,-0.85\myheight)--(A2.180);
		\draw [copy] (pic2.0)--(0.45\myheight,-1.1\myheight)--(0.45\myheight,-1.35\myheight)--(R2.180);
		
		\draw [copy] (pic4.0)--(0.45\myheight,-2.2\myheight)--(0.45\myheight,-1.95\myheight)--(A4.180);
		\draw [copy] (pic4.0)--(0.45\myheight,-2.2\myheight)--(0.45\myheight,-2.45\myheight)--(R4.180);
		
		\node (U1) at (1.7\myheight,-1.1\myheight) [rotate=90,draw,rectangle,rounded corners,minimum width=1\myheight] {\tiny Deconv $(2)$};
		\node (U2) at (1.7\myheight,-2.2\myheight) [rotate=90,draw,rectangle,rounded corners,minimum width=1\myheight] {\tiny  Deconv $(4)$};
		
		\node (S1) at (2.1\myheight,-1.1\myheight) [rotate=90,draw,rectangle,rounded corners] {\tiny Softmax};

		\draw [ru] (R2-|U1.south)--(1.9\myheight,-1.35\myheight)--(1.9\myheight,-1.65\myheight)--(3.6\myheight,-1.65\myheight)--(m2);
		\draw [ru] (R1)--(1.9\myheight,-0.25\myheight)--(1.9\myheight,-0.55\myheight)--(3.6\myheight,-0.55\myheight)--(m1);
		\draw [ru] (R4-|U2.south)--(1.9\myheight,-2.45\myheight)--(1.9\myheight,-2.75\myheight)--(3.6\myheight,-2.75\myheight)--(m4);
		
		\draw [att] (A4.0)--(A4-|U1.north);
		\draw [ru] (R4.0)--(R4-|U1.north);
		\draw [att] (A2.0)--(A2-|U1.north);
		\draw [ru] (R2.0)--(R2-|U1.north);
		
		\draw [att] (A2-|U1.south)--(1.9\myheight,-0.85\myheight)--(1.9\myheight,-1.1\myheight)--(S1);
		
		\draw [att] (A4-|U1.south)--(S1|-A4)--(S1);
		\draw [att] (A1.0)--(S1|-A1)--(S1);
		
		\draw [att] (S1)--(2.4\myheight,-1.1\myheight)--(2.4\myheight,0)--(att1);
		\draw [att] (S1)--(att2);
		\draw [att] (S1)--(2.4\myheight,-1.1\myheight)--(2.4\myheight,-2.2\myheight)--(att4);
		
		\draw [att] (att1)--(m1);
		\draw [att] (att2)--(m2);
		\draw [att] (att4)--(m4);			
	\end{tikzpicture}
}
\begin{figure*}[ht]
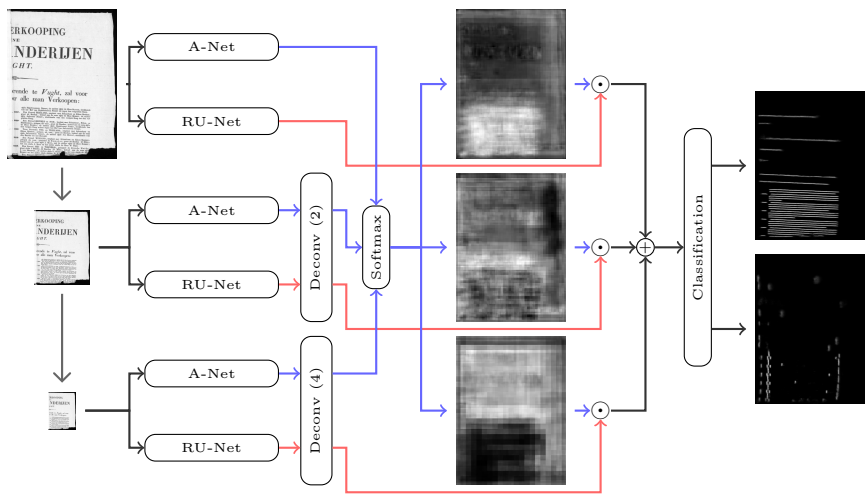

	\centering
	\drawARU
	\caption{\textbf{ARU-Net --} The input image and its downscaled versions are fed into the A-Net and R-U-Net (weight sharing accross different scales). 
	The results for the lower resolutions are deconvolved. The attention maps are passed through a softmax normalization. The brighter the map at a certain position the more attention is paid to that position at the corresponding scale. 
	The attention maps are point-wise muliplied with the feature maps of the RU-Net. 
	The results are summed and a classification is performed.}
	\label{fig:arunet}
\end{figure*}
\begin{defi}[ARU-Net]
 An RU-Net incorporating the described spatial attention mechanism is called \textit{ARU-Net}, see Fig.~\ref{fig:arunet}.
\end{defi}
The point-wise multiplication combined with the pixel-wise attention maps allow the ARU-Net to pay attention in different scales at different positions of the image. In Fig.~\ref{fig:arunet}
one can see that this behavior was indeed learned by the network. It seems like the RU-Net is specialized on a certain font size and the A-Net distinguishes between areas of different font sizes (bright and dark areas).

The ARU-Net as introduced can be used for any pixel labeling task, e.g., binarization, page detection and page segmentation. The purpose of the ARU-Net is defined and fixed by the number 
of classes and the ground truth data provided for training. In this work, we limit ourselves to the baseline detection problem introduced in Sec.~\ref{ssec:ps}. 
For this purpose, we introduce three different classes: baseline (bl), separator (sep) and other ($\varnothing$). The separators mark beginning and end of each text line. 
Although, the separator information is implicitly encoded by the baselines, it is advantageous to explicitly introduce it as possible classification result. 
Especially, for baselines which are close together, e.g., such belonging to two adjacent columns, this approach helps to avoid segmentation errors.
Pixel ground truth for the classes $\mathcal{C}=\{\text{bl},\text{sep},\varnothing\}$ could be automatically generated by Alg.~S.1 (supplements) given the baseline ground truth. 

\begin{figure}[ht]
	\centering
	\subfloat[\textbf{Baseline ground truth --} The baselines are defined by the red dots. Dots of the same baseline are connected.]{\includegraphics[width=0.475\textwidth]{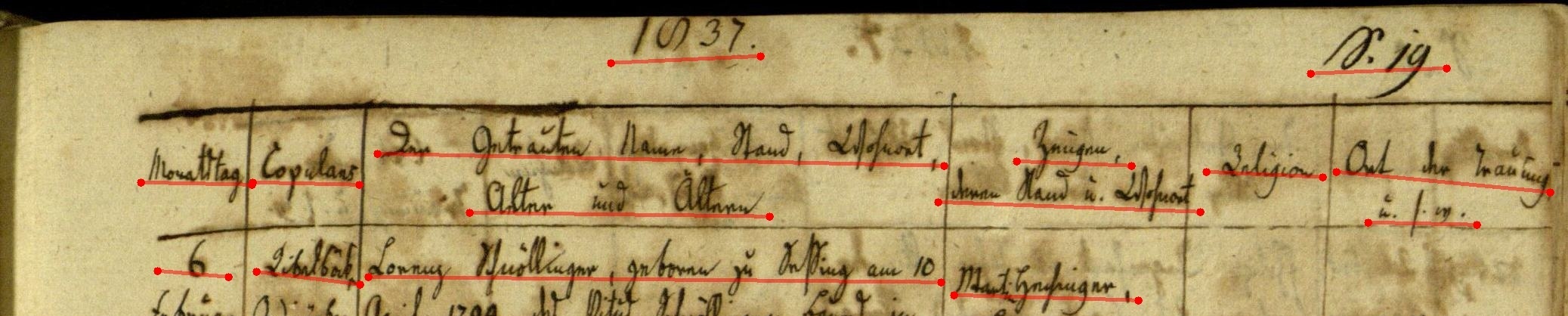}
		\label{fig:gt1}}
	\hfil
	\subfloat[\textbf{Pixel ground truth --} Green encodes the separator class, red the baseline class and black the ''other'' class. See supplements for an algorithm to automatically generate such GT.] {\includegraphics[width=0.475\textwidth]{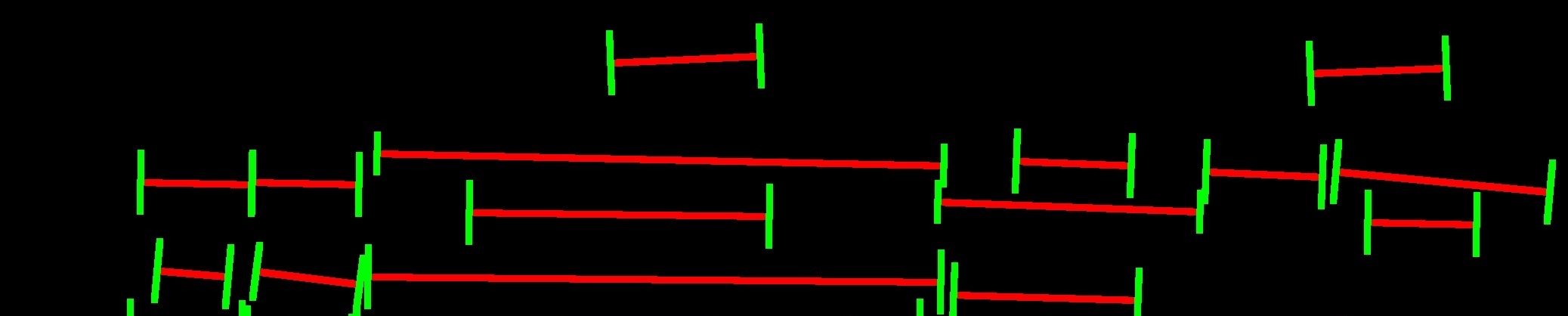}
		\label{fig:gt2}}
	\caption{\textbf{Baseline and pixel ground truth --} These are shown for the top snippet of the image of Fig.~\ref{fig:wf}.}
	\label{fig:gt}
\end{figure}

\begin{figure}[ht]
	\centering
	\subfloat[\textbf{ARU-Net output --} The estimated baselines $B$ (blue) and separators $S$ (cyan) are shown.]{\includegraphics[width=0.475\textwidth]{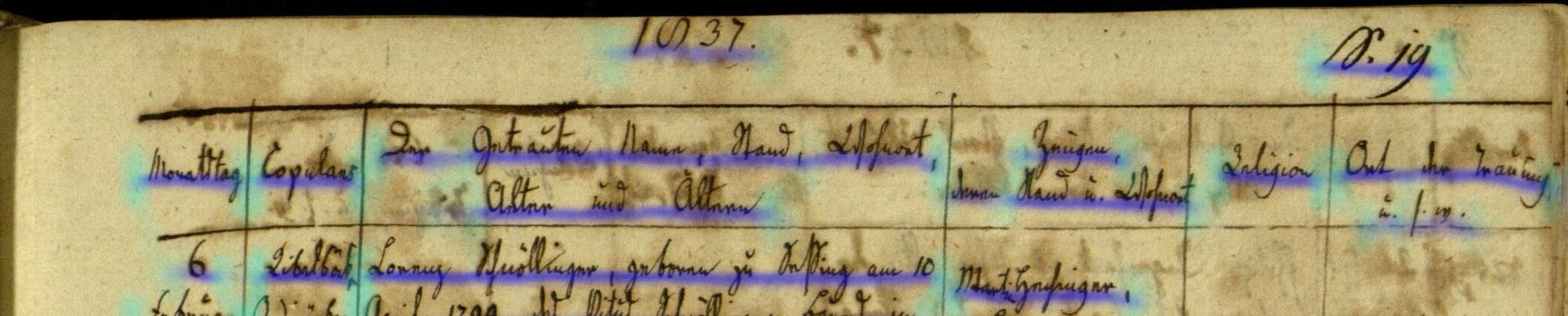}
		\label{fig:aruout}}
	\hfil
	\subfloat[\textbf{Superpixel and neighborhood system --} The calculated SPs (blue) are shown along with the resulting Delaunay neighborhood system $\mathcal{N}$ (yellow).]{\includegraphics[width=0.475\textwidth]{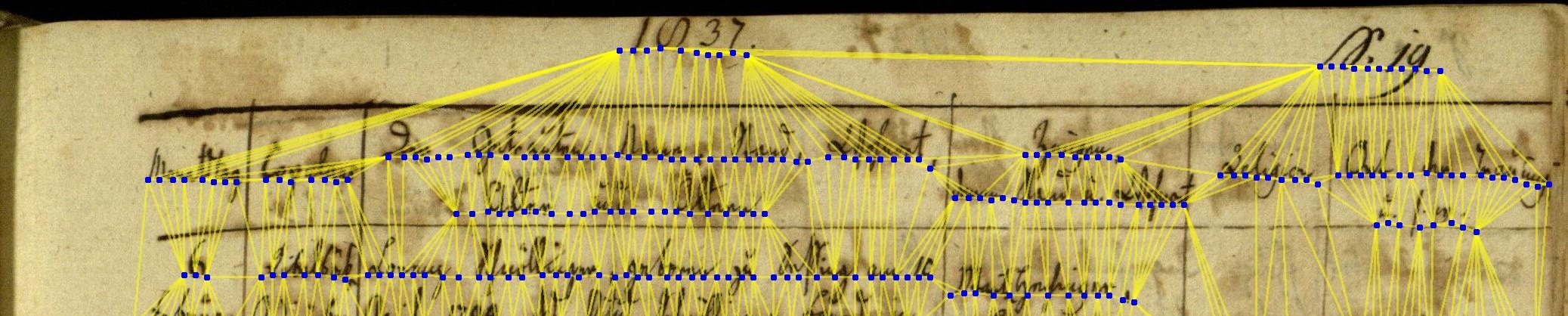}
		\label{fig:spnei}}
	\caption{\textbf{Baseline detection process -- } Two intermediate steps are shown for the top snippet of the image of Fig.~\ref{fig:wf}.}
	\label{fig:spcalc}
\end{figure}

A sample image with baseline ground truth along with its generated pixel ground truth is depicted in Fig.~\ref{fig:gt}. The prediction of a trained ARU-Net for this sample image is shown in Fig.~\ref{fig:aruout}.
\subsection{Stage II: Baseline Estimation}
This subsection describes the second stage of the proposed approach. Baselines are estimated given the output of the ARU-Net.
This task consists of three steps: superpixel calculation, state estimation and superpixel clustering, which are described in the following.

The trained ARU-Net generates an output $C\in\left[0,1\right]^{I_h\times I_w\times 3}$ for each image $I\in\mathcal{I}$. In the following
$B=C_{:,:,1}$ denotes the image encoding the confidence of each pixel belonging to a baseline and $S=C_{:,:,2}$ is the separator image, see Fig.\ref{fig:aruout}
\subsubsection{Superpixel Calculation}
The number of all pixels in an image often exceeds several millions. To reduce the dimensionality of the problem (the number of pixels to be regarded for the baseline estimation), we limit ourselves to a subset of all pixels.
\begin{defi}[superpixel]
 Let $\mathcal{S}=\{\boldsymbol{p}_1,...,\boldsymbol{p}_N\}$ be a subset of the image pixels of $I$ (typically, $N \ll I_h\cdot I_w$ holds). An element of $\mathcal{S}$ is called \textit{superpixel (SP)}.
\end{defi}
Basically, the definition of a superpixel does not introduce any new concept. A SP is just a normal pixel which is somehow regarded to be of certain importance.
Since it is a frequently used term, we decided to introduce it via a definition.
It is easy to see that the choice of the set of SPs is crucial for the overall performance. If there are no SPs for a baseline at all, this baseline will be missed.
To calculate a suitable set of SPs, we utilize the baseline map $B$ generated by the ARU-Net.

In a first step $B$ is binarized $B_b = B>b$ by an element-wise comparison of $B$ with a confidence threshold $b$. The morphological skeleton $B_s=\text{SKE}(B_b)$ is calculated for $B_b$ following Lantu\'{e}joul's formula \cite{Serra1982}. 
All foreground pixels (pixels with an intensity of $1$) of $B_s$ build an initial set of pixels $\left\{\boldsymbol{p}_1,...,\boldsymbol{p}_M\right\}$. Its elements are sorted in descending order w.r.t. their baseline confidences.
Finally, $\mathcal{S}$ is set up by iteratively adding pixels of this sorted list(beginning with the first pixel). To keep the number of SPs small, a new pixel $\boldsymbol{p}$ is added to $\mathcal{S}$ only if its distance to all other SPs exceeds a certain threshold $\left\|\boldsymbol{p}-\boldsymbol{q}\right\|_2>d\quad\forall \boldsymbol{q}\in\mathcal{S}$, otherwise it is skipped. In Fig.~\ref{fig:spnei} the set of resulting SPs is shown. These SPs build the basis for the further clustering.
\begin{rem}
	For all experiments, we have chosen fixed values of $b=0.2$ (binarization threshold) and $d=10$ (distance threshold).
	These demonstrated to be well suited for a wide range of different scenarios. Hence, they are not regarded as free parameters of the system 
	which have to be further tuned. This also holds for the parameters which are fixed in Rem.~\ref{rem:p2} \& \ref{rem:fe}.
\end{rem}
\subsubsection{Superpixel State Estimation}
\label{ssec:se}
Assume we can assign each SP to a certain text line. The state of an SP should encode meaningful characteristics of its text line. These characteristics will be defined and combined to build the state.
This work is based on previous work of \cite{Ryu2014,gr2017}, but adapted to the characteristics of SPs extracted given the ARU-Net output, e.g., easier calculation of the local text orientation as well as a different smoothing cost formulation.
\begin{defi}[local text orientation]
\label{def:lo}
 The \textit{local text orientation} $\theta$ of an SP $\boldsymbol{p}$ is the slope of its text line's baseline at the coordinates closest (w.r.t. the euclidean distance) to $\boldsymbol{p}$.  
\end{defi}
\begin{defi}[interline distance]
\label{def:id}
 The \textit{interline distance} $s$ of an SP $\boldsymbol{p}$ is the distance of its text line's baseline to the nearest other baseline. 
 Distance means the distance which is orthogonal to the local text direction of $\boldsymbol{p}$.
\end{defi}
\begin{defi}[state]
\label{def:state}
 The state of an SP is the pair $(\theta,s)$ of its local text orientation and its interline distance.
\end{defi}
In the following, we will describe a method to estimate the states of all SPs. The local text orientation will be calculated in a straightforward way utilizing solely the baseline image $B$ and local information.
On the other hand, the estimation of the interline distances combines local information of the text line's periodicity with the more global assumption that nearby SPs tend to have similar interline distances.
For these approaches the concepts of neighborhood and connectivity are mandatory and will be introduced.
\begin{defi}[neighborhood system, edge, adjacent]
 We call a subset $\mathcal{N}\subset \mathcal{S}\times\mathcal{S}$ \textit{neighborhood system}. An element of 
$\mathcal{N}$ is called \textit{edge} and denoted by $\boldsymbol{e}_{\boldsymbol{p},\boldsymbol{q}}$. 
$\mathcal{N}$ is not directed ($\boldsymbol{e}_{\boldsymbol{p},\boldsymbol{q}}=\boldsymbol{e}_{\boldsymbol{q},\boldsymbol{p}}$).
Two SPs $\boldsymbol{p},\ \boldsymbol{q}$ are \textit{adjacent} if $\boldsymbol{e}_{\boldsymbol{p},\boldsymbol{q}}\in\mathcal{N}$.
$\boldsymbol{e}_{\boldsymbol{p},\boldsymbol{q}}\setminus \boldsymbol{p}\in\mathcal{S}$ denotes the SP $\boldsymbol{q}$.
\end{defi}
\begin{rem}
 In the following the neighborhood system $\mathcal{N}$ for a set of SPs is always calculated by Delaunay's triangulation \cite{Delaunay1934}. 
\end{rem}
\begin{defi}[connectivity function]
\label{def:cof}
 The line segment $g(\ \cdot\ ;\boldsymbol{e}_{\boldsymbol{p},\boldsymbol{q}}):\left[0,1\right]\rightarrow\mathbb{R}^2$ defined by $g(\tau; \boldsymbol{e}_{\boldsymbol{p},\boldsymbol{q}})\coloneqq\boldsymbol{p} + \tau \left(\boldsymbol{q}-\boldsymbol{p}\right)$ connects the two pixels $\boldsymbol{p},\boldsymbol{q}$ of the edge $\boldsymbol{e}_{\boldsymbol{p},\boldsymbol{q}}$. The function $\Gamma : \mathcal{N}\times \mathcal{I}\rightarrow \left[0,1\right]$ defined by
 \begin{align*}
 \Gamma(\boldsymbol{e}_{\boldsymbol{p},\boldsymbol{q}},I)=\frac{\int_{0}^{1} I(g(\tau;\boldsymbol{e}_{\boldsymbol{p},\boldsymbol{q}})) d\tau}{\left\|\boldsymbol{p}-\boldsymbol{q}\right\|_2}
\end{align*}
is called \textit{connectivity function}. $I(g(\tau;\boldsymbol{e}_{\boldsymbol{p},\boldsymbol{q}}))$ denotes the intensity of the pixel in $I$ closest (w.r.t the euclidean distance) to the real-valued coordinates $g(\tau;\boldsymbol{e}_{\boldsymbol{p},\boldsymbol{q}})$.
\end{defi}
The connectivity function calculates the average intensity for a given image along the shortest path connecting two pixels.
The local text orientation of each SP is estimated by $\theta_{\boldsymbol{p}}=\text{LTO}(\boldsymbol{p};\mathcal{N},B)$ utilizing $\mathcal{N}$ and the baseline image $B$, see Alg.~\ref{alg:lo}. The LTO algorithm picks the two neighbors of an SP $\boldsymbol{p}$ with the largest baseline connectivity to $\boldsymbol{p}$ and determines the slope of the line passing through these neighbors.
\begin{algorithm}
\DontPrintSemicolon
\SetKwComment{Comment}{$\rhd$ }{}
\SetKwInOut{Input}{input}\SetKwInOut{Output}{output}\SetKwInOut{Return}{return}
\vspace{1mm}
\Input{SP $\boldsymbol{p}$, neighborhood system $\mathcal{N}$, baseline image $B$}
\Output{local text orientation $\theta$ of $\boldsymbol{p}$}
$\mathcal{M} \gets \left\{\boldsymbol{e}_{\boldsymbol{q},\boldsymbol{r}}\in\mathcal{N}\ |\ \boldsymbol{q}=\boldsymbol{p} \lor \boldsymbol{r}=\boldsymbol{p}\right\}$\\
$\boldsymbol{L} \gets $ sorted list of $\mathcal{M}$ \Comment*{sorted by $\Gamma(\boldsymbol{e}_{\boldsymbol{q},\boldsymbol{r}},B)$}
\uIf{$\left|\boldsymbol{L}\right| == 1$}{
$\boldsymbol{e}_{\boldsymbol{q},\boldsymbol{r}} \gets \boldsymbol{L}_1$\Comment*{$\boldsymbol{L}_k$ is the $k$-th element}
}
\Else{
$\boldsymbol{e}_{\boldsymbol{q},\boldsymbol{r}} \gets \left(\boldsymbol{L}_1\setminus \boldsymbol{p},\boldsymbol{L}_2\setminus \boldsymbol{p}\right)$
}
\Return{$\theta \gets \arctan\left(\frac{\boldsymbol{r}_y-\boldsymbol{q}_y}{\boldsymbol{r}_x-\boldsymbol{q}_x}\right)$}
\vspace{1mm}
\caption{Local Text Orientation of $\boldsymbol{p}$}\label{alg:lo}
\end{algorithm}

\begin{figure*}
	\centering
	\includegraphics[width=0.95\textwidth]{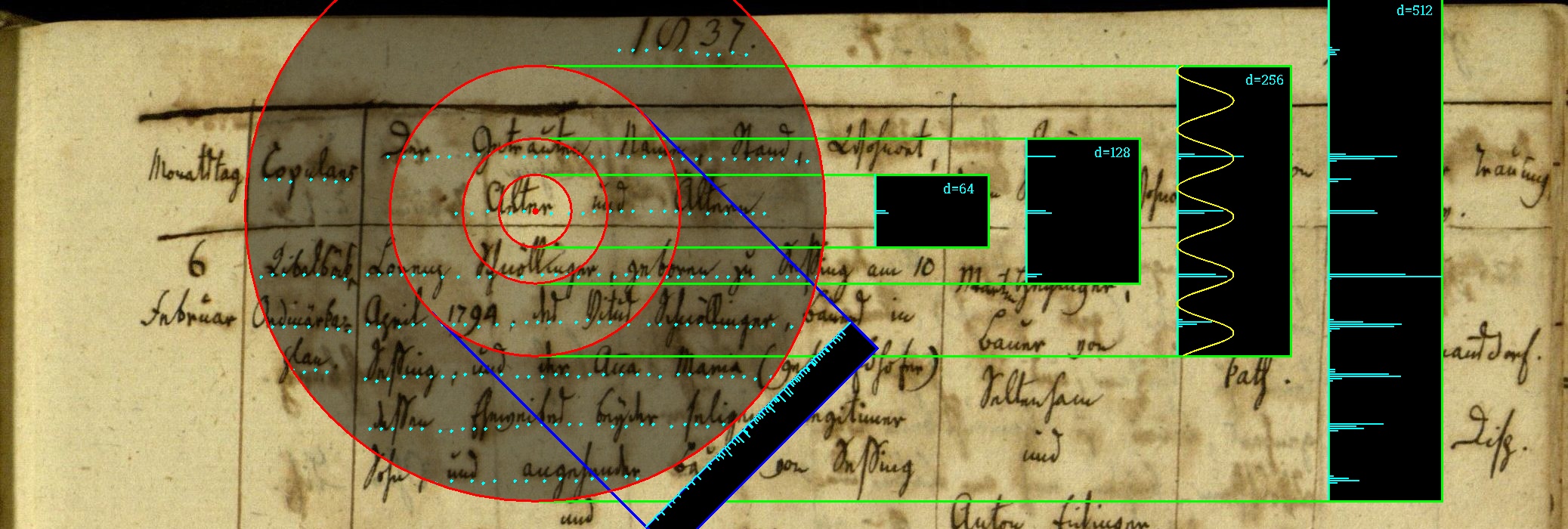}
	\caption{\textbf{Interline distance estimation --} Illustration of several projection profiles for a certain SP (red point). The profiles for different diameters $d\in\{64,128,256,512\}$ and an orientation of $0\degree$ are shown in green. 
		The winning period (interline distance) is drawn as yellow curve. In blue a histogram for a wrong orientation ($45\degree$) is shown.}
	\label{fig:pp}
\end{figure*}

The periodicity of text lines in document images is utilized to calculate the interline distances. We determine the interline distance of an SP $\boldsymbol{p}$ by evaluating the regional
text-line periodicity around $\boldsymbol{p}$ as follows. For an SP $\boldsymbol{p}$, a circular region of diameter $d\in\mathbb{N}$ around $\boldsymbol{p}$, and a projection direction determined by the local text orientation $\theta_{\boldsymbol{p}}$, let 
$\boldsymbol{h}^{\boldsymbol{p},d}=(h^{\boldsymbol{p},d}_1,...,h^{\boldsymbol{p},d}_{d})\in\mathbb{N}^d$
be the projection profile with respect to $\mathcal{S}$, see Fig.~\ref{fig:pp}. 
For the calculation of $\boldsymbol{h}^{\boldsymbol{p},d}$, only SPs with a distance to $\boldsymbol{p}$ of less than $\frac{d}{2}$ are taken into account.
\begin{rem}
 The projection profile $\boldsymbol{h}^{\boldsymbol{p},d}$ can be calculated very efficiently by utilizing the cross product of the orientation vector $\boldsymbol{o}=\left(\cos (\theta_{\boldsymbol{p}}),\sin(\theta_{\boldsymbol{p}})\right)^T$ and 
 the vectors $\vv{\boldsymbol{pq}}$ for $\boldsymbol{q}\in\mathcal{S}$ with $\left\|\boldsymbol{p}-\boldsymbol{q}\right\|_2\leq\frac{d}{2}$.
\end{rem}
To extract the regional periodicity inherent in the projection profile $\boldsymbol{h}^{\boldsymbol{p},d}$, a Discrete Fourier Transformation (DFT) is applied to $\boldsymbol{h}^{\boldsymbol{p},d}$ with resulting coefficients
$\boldsymbol{H}^{\boldsymbol{p},d}=\left(H^{\boldsymbol{p},d}_1,...,H^{\boldsymbol{p},d}_{d}\right)\in\mathbb{C}^d$.
A coefficient $H^{\boldsymbol{p},d}_k,\ k\in\left\{1,...,d\right\}$ corresponds to the portion of the signal with a period of $\frac{d}{k}$ to the entire signal $\boldsymbol{h}^{\boldsymbol{p},d}$. 
In the simplest case, the index $k'$ of the dominant coefficient of $\boldsymbol{H}^{\boldsymbol{p},d}$ determines
the interline distance $s$ of $\boldsymbol{p}$ as $s = \frac{d}{k'}$. However, we may be forced
to assign a different value to $\boldsymbol{s}$ due to additional constraints to be discussed in a moment. 
Therefore, we introduce a data energy value for each possible value $\frac{d}{k}$ of the interline distance $s$ of $\boldsymbol{p}$. From
energy, we then derive a data cost to be used within a cost minimization framework for finding the optimal interline distance. 
\begin{defi}[data energy, data cost]
 The \textit{data energy} of SP $\boldsymbol{p}$ and interline distance $\frac{d}{k}$ is given by
$E_{\boldsymbol{p}}\left(\frac{d}{k}\right)=\frac{\left|H^{\boldsymbol{p},d}_k\right|^2}{\left\|\boldsymbol{H}^{\boldsymbol{p},d}\right\|_2^2}$.
The \textit{data cost} is calculated by
$D_{\boldsymbol{p}}\left(\frac{d}{k}\right) = -\log \left(E_{\boldsymbol{p}}\left(\frac{d}{k}\right) \right)$.
\end{defi}
Remarkably, the data energy is normalized such that it sums (over $k$) up to $1.0$ for arbitrary $d\in\mathbb{N}$. 
To cover a suitable range of different interline distances as well as to be robust against disturbances due to close-by text regions of a different style, the projection profiles and 
DFTs are calculated for different diameters $d\in\left\{64,128,256,512\right\}$ and $k\in\left\{3,4,5\right\}$.
The choice of the values for $d$ and $k$ is application driven and results in reasonable interline distances (sorted list) of $\boldsymbol{S}\coloneqq$(170.7, 128.0, 102.4, 85.3, 64.0, 51.2, 42.7, 32.0, 25.6, 21.3, 16.0, 12.8).

\begin{figure}
	\centering
	\subfloat[\textbf{Greedy states --} The SP states for a greedy labeling using the highest data energy are shown.]{\includegraphics[width=0.475\textwidth]{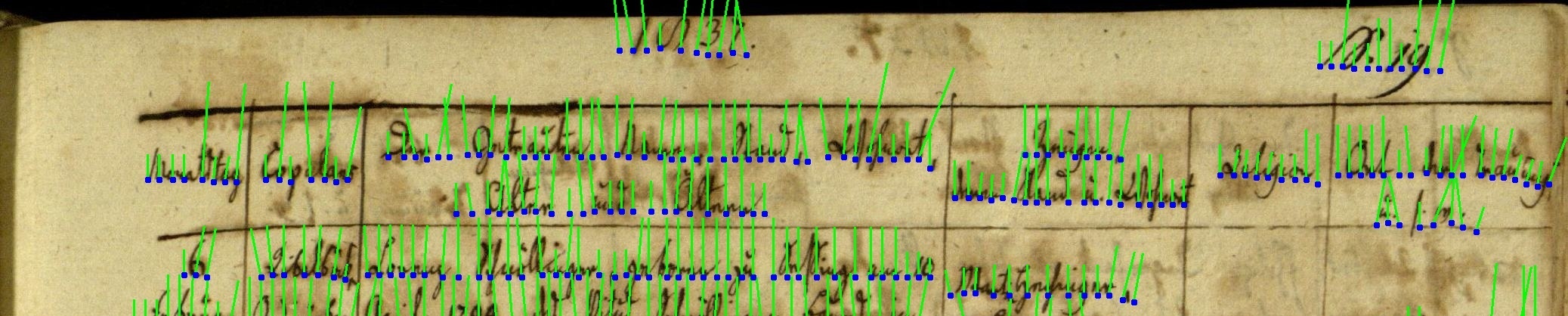}
		\label{fig:stgr}}
	\hfil
	\subfloat[\textbf{Smoothed states --} The SP states for a cost optimal labeling are shown.]{\includegraphics[width=0.475\textwidth]{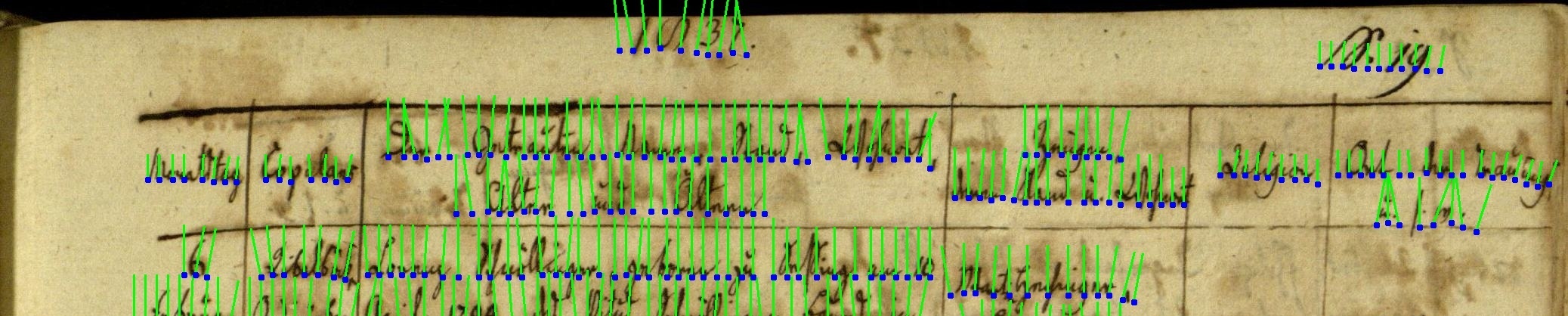}
		\label{fig:st}}
	\caption{\textbf{SPs with their assinged states --} The local text orientation of each SP is visualized by the orientation of the green lines (rotated by $90\degree$). 
		The length of the lines encode the interline distance of the corresponding SP.}
	\label{fig:spst}
\end{figure}

In the following, we write $s_{\boldsymbol{p}}$ for the assigned interline distance $s=\frac{d}{k}\in\boldsymbol{S}$ of SP $\boldsymbol{p}$ and say $\boldsymbol{p}$ is labeled with $s_{\boldsymbol{p}}$.
A labeling $\left\{ s_{\boldsymbol{p}} \right\}_{{\boldsymbol{p}}\in\mathcal{S}}$ of $\mathcal{S}$ assigns an interline distance to each SP of $\mathcal{S}$.
Following a greedy labeling strategy by assigning the interline distance with the highest data energy to each SP leads to a noisy result, see Fig.~\ref{fig:stgr}. 
To reduce the noise effects, the influence of close-by SPs is taken into account. It is reasonable to expect that neighboring SPs tend to have similar interline distances. This expectation is
encoded via a smoothing cost defined for adjacent SPs.
\begin{defi}[smoothing cost]
 For each $\boldsymbol{e}_{\boldsymbol{p},\boldsymbol{q}}\in\mathcal{N}$ (and assigned interline distances $s_{\boldsymbol{p}},s_{\boldsymbol{q}}$) the \textit{smoothing cost} is defined by
\begin{align*}
V_{\boldsymbol{p},\boldsymbol{q}}(s_{\boldsymbol{p}},s_{\boldsymbol{q}}) = \begin{cases}
                                                                            \sigma & ,\ \langle s_{\boldsymbol{p}},s_{\boldsymbol{q}}\rangle_s \geq 4,\\
                                                                            \langle s_{\boldsymbol{p}},s_{\boldsymbol{q}}\rangle_s & ,\ else.\\
                                                                           \end{cases}
\end{align*}
\end{defi}
$\langle s_{\boldsymbol{p}},s_{\boldsymbol{q}} \rangle_s$ is the index difference of $s_{\boldsymbol{p}}$ and $s_{\boldsymbol{q}}$ in the sorted list $\boldsymbol{S}$ of possible interline distances, e.g., $\langle 16.0, 42.7\rangle_s=4$.
Thus, the smoothing cost $V_{\boldsymbol{p},\boldsymbol{q}}(s_{\boldsymbol{p}},s_{\boldsymbol{q}})$ becomes large if interline distances of different size are assigned to adjacent SPs. 
A maximum cost value of $\sigma$ is used for huge differences in the interline distances. Setting $\sigma$ to a large value prevents neighboring SPs to differ to much in their interline distances.
\begin{defi}[labeling cost]
 The \textit{labeling cost} is given by
\begin{align*}
 C\left(\left\{ s_{\boldsymbol{p}} \right\}_{{\boldsymbol{p}}\in\mathcal{S}}\right) = \alpha\sum_{{\boldsymbol{p}}\in\mathcal{S}} D_{\boldsymbol{p}}\left(s_{\boldsymbol{p}}\right)
  +\beta\sum_{\boldsymbol{e}_{\boldsymbol{p},\boldsymbol{q}}\in \mathcal{N}}V_{\boldsymbol{p},\boldsymbol{q}}(s_{\boldsymbol{p}},s_{\boldsymbol{q}}).
\end{align*}
\end{defi}
The data cost and the smoothing costs are weighted by $\alpha$ and $\beta$, respectively, to form the labeling cost.
The graphcut algorithm \cite{Boykov2001} is utilized to minimize the labeling cost. 
The final labeling is shown in Fig.~\ref{fig:st}. 
\begin{rem}
	\label{rem:p2}
	For all experiments, we have chosen fixed values of $\sigma=25$, $\alpha=1$ and $\beta=1$.
\end{rem}
\subsubsection{Superpixel Clustering}
\label{sssec:sc}
In the previous subsections the calculation of SPs and their enrichment with state information was described.
In a final step, this state information is utilized to cluster the SPs to build baselines. There will be a one-to-one assignment between clusters and baselines.
In the following, we call a set of SPs \textit{cluster}.

\begin{figure}[ht]
	\centering
	\subfloat[\textbf{Without separator information -- } The entire neighborhood system (yellow) is shown.]{\includegraphics[width=0.475\textwidth]{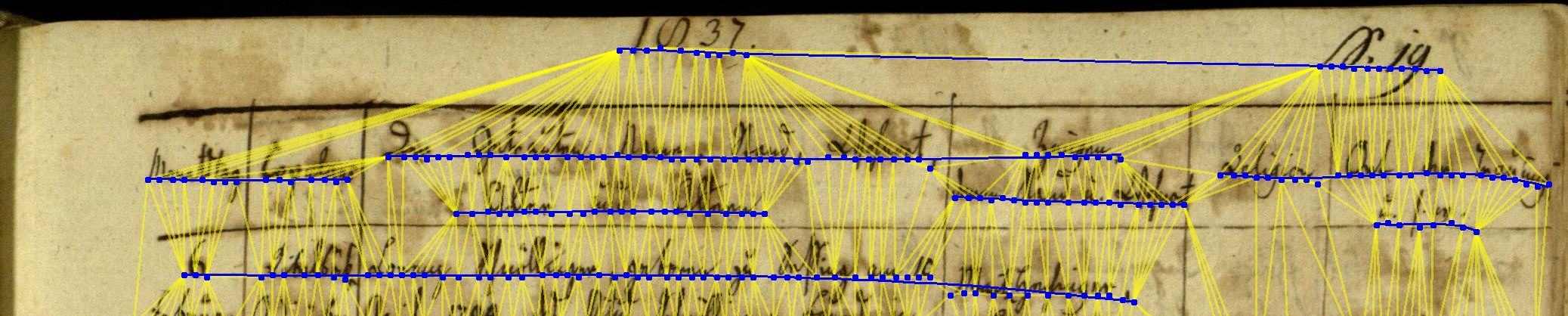}
		\label{fig:finA}}
	\hfil
	\subfloat[\textbf{With separator information -- } The neighborhood system was reduced by removing edges (cyan) with high separator connectivity. The corresponding separator information is illustrated in Fig.~\ref{fig:aruout}.]{\includegraphics[width=0.475\textwidth]{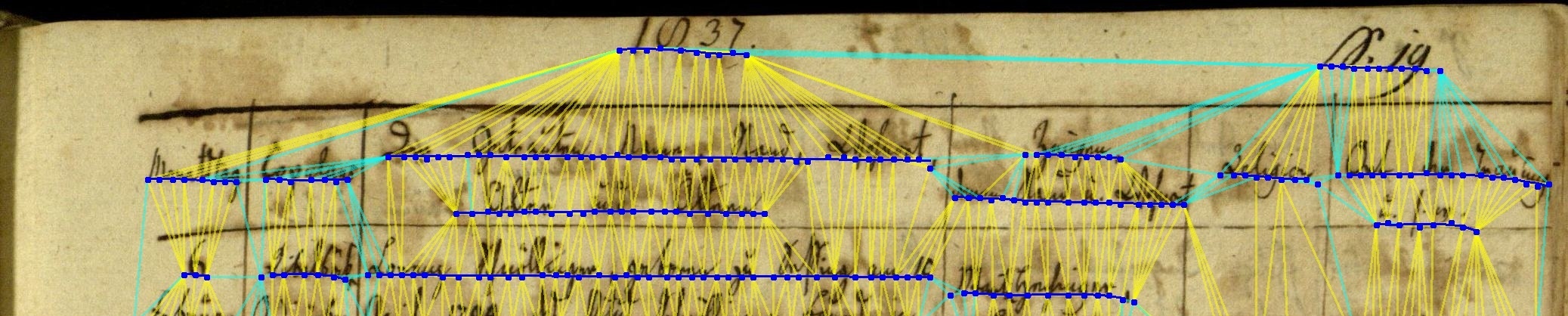}
		\label{fig:finB}}
	\caption{\textbf{Influence of the separator information --} The resulting baselines (blue lines) with and without taking into account the separator information are shown.}
	\label{fig:fin}
\end{figure}

In this subsection we formulate the clustering problem and introduce a greedy clustering procedure to solve the problem.
Two assumptions which hold for baselines in general constitute the conditions for the clustering problem:
\begin{enumerate}[label=(\Roman*)]
\item Baselines should not exceed a certain curvilinearity value.\label{cond:I}
\item Within the interline distance of a baseline there are no other baselines.\label{cond:II}
\end{enumerate}
Basically, assumption \ref{cond:I} claims that a baseline can be approximated by a polynomial function of a certain degree, see \cite{Ryu2014}. Assumption \ref{cond:II} is self-explanatory. 
\begin{rem}
	In the following, $\theta(\{\boldsymbol{p}_1,...,\boldsymbol{p}_n\})$ denotes the average orientation and $s(\{\boldsymbol{p}_1,...,\boldsymbol{p}_n\})$ the average interline distance of all SPs in 
	$\{\boldsymbol{p}_1,...,\boldsymbol{p}_n\}$.
\end{rem}
\begin{defi}[curvilinearity value]
\label{def:rp}
Let $deg\in\mathbb{N}$ and $\mathcal{S}$ be a set of SPs.
 Assume $p_{\mathcal{S},deg}(t)\in\mathbb{P}\left[t\right]$ is the polynomial which solves the linear regression problem in the monomials $t^0,t^1,...,t^{deg}$ for $\mathcal{S}'$ which results from $\mathcal{S}$ by rotating all pixels by $-\theta(\mathcal{S})$. 
 The root-mean-square regression error normalized by $s(\mathcal{S})$ is called \textit{curvilinearity value} of $\mathcal{S}$ and is denoted by $cur(\mathcal{S},deg)$.
\end{defi}
\begin{rem}
We fix $deg=3$ and omit it in the following.
\end{rem}
Def.~\ref{def:rp} allows for an easy evaluation of \ref{cond:I}. To test for \ref{cond:II} we will introduce the distance of two clusters.
Remarkably, only distances orthogonal to the text orientation should be taken into account. First, the orthogonal component of the distance between two SPs is introduced.
Afterwards, this is generalized for two clusters of SPs.
\begin{defi}[off-text distance]
\label{def:otd}
 Given two SPs $\boldsymbol{p}$, $\boldsymbol{q}$ and an orientation $\theta$, the \textit{off-text distance} of $\boldsymbol{p}$ and $\boldsymbol{q}$ is the length of the component of $\boldsymbol{p}-\boldsymbol{q}\in\mathbb{R}^2$ 
 which is orthogonal to $\theta$. It is denoted by $\left\|\boldsymbol{p}-\boldsymbol{q} \right\|_{\theta}$.
\end{defi}
\begin{rem}
The off-text distance can be efficiently calculated by
$
  \left\|\boldsymbol{p}-\boldsymbol{q} \right\|_{\theta}= \left|\left( \boldsymbol{p}_x - \boldsymbol{q}_x\right)\sin (\theta)-\left( \boldsymbol{p}_y - \boldsymbol{q}_y\right)\cos (\theta) \right|$.
\end{rem}
Calculating the minimal pairwise off-text distance of all SPs of two clusters could result in a cluster distance distorted by SP outliers. Therefore, SPs in each cluster will be projected onto the corresponding regression curve obtained by the regression problem in
Def.~\ref{def:rp}, before taking pairwise distances. 
\begin{defi}[regression curve]
 Let $\mathcal{S},\ \mathcal{S}'$ and $p_{\mathcal{S}}(t)$ be of Def.~\ref{def:rp}. The spatial t-range of $\mathcal{S}'$ is given by $t_{min}=\min\{\boldsymbol{p}_x\ \vert\ \boldsymbol{p}\in\mathcal{S}'\}$
 and $t_{max}=\max\{\boldsymbol{p}_x\ \vert\ \boldsymbol{p}\in\mathcal{S}'\}$.
 A curve $c_{\mathcal{S}}:\ \left[0,1\right]\rightarrow \mathbb{R}^2$ which results from rotating the graph of $p_{\mathcal{S}}(t)$ for $t\in\left[t_{min},t_{max}\right]$ by $\theta(\mathcal{S})$
 is called \textit{regression curve} of $\mathcal{S}$.
\end{defi}
The SPs in $\mathcal{S}$ are projected onto $c_{\mathcal{S}}$ (in direction $\theta(\mathcal{S})+\frac{\pi}{2}$). The resulting projected SPs are denoted by $\mathcal{S}^c$.
To achieve robust distance estimates even for curved and differently slanted text lines we focus on SPs of the different clusters which are quite close to each other and furthermore take into account the slope 
of the regression curve at the specific SP positions instead of averaging over the entire text line.
\begin{defi}[cluster distance]
 Assume two clusters $\mathcal{S}_1$, $\mathcal{S}_2$ with regression curves $c_{\mathcal{S}_1}(t)$, $c_{\mathcal{S}_2}(t)$ and projected SPs $\mathcal{S}^c_1$, $\mathcal{S}^c_2$.
  The \textit{cluster distance} is defined as 
 \begin{align*}
  d(\mathcal{S}_1,\mathcal{S}_2)=\min_{\substack{\boldsymbol{p}\in\mathcal{S}_1^c, \boldsymbol{q}\in\mathcal{S}_2^c: \\ \left\|\boldsymbol{p}-\boldsymbol{q}\right\|_2<4\cdot s(\mathcal{S}_1\cup\mathcal{S}_2)}}
   \left\|\boldsymbol{p}-\boldsymbol{q} \right\|_{\theta^c(\boldsymbol{p},\boldsymbol{q})},
 \end{align*}
where $\theta^c(\boldsymbol{p},\boldsymbol{q})$ is the average slope of the corresponding regression curves at $\boldsymbol{p}$ and $\boldsymbol{q}$, respectively.
\end{defi}
Since, it is now possible to evaluate conditions \ref{cond:I} \& \ref{cond:II}, we will use this to introduce feasible sets of clusters.  
For this purpose, we will limit ourselves to partitions (a special kind of cluster sets) and require the baseline clusters to be $\mathcal{N}$-linked. The set of all partitions of a set $\mathcal{M}$ is denoted by $par(\mathcal{M})$.
\begin{defi}[$\mathcal{N}$-linked]
 Let $\mathcal{S}$ be a cluster and $\mathcal{N}$ be a neighborhood system. $\mathcal{S}$ is $\mathcal{N}$\textit{-linked} iff $\forall \boldsymbol{p},\boldsymbol{q}\in\mathcal{S}\ \exists \boldsymbol{p}_0,...,\boldsymbol{p}_N\in\mathcal{S}: \boldsymbol{p}_0=\boldsymbol{p}\land\boldsymbol{p}_N=\boldsymbol{q}\land\boldsymbol{e}_{\boldsymbol{p}_i,\boldsymbol{p}_{i+1}}\in\mathcal{N}\ (0\leq i\leq N-1)$
holds.
\end{defi}
That means, for all pairs of SPs there are edges in $\mathcal{S}$ which connect these respective SPs.
\begin{defi}[feasible]
\label{def:fe}
 For $\gamma,\delta\in\mathbb{R}_+$, $L\in\mathbb{N}$, a set of SPs $\mathcal{S}$ and a neighborhood system $\mathcal{N}$, we call a set of clusters $\mathscr{P}=\left\{\mathcal{S}_0, ... , \mathcal{S}_L \right\}$ \textit{feasible} 
 iff 
 \begin{enumerate}
  \item $\mathscr{P}\in par(\mathcal{S})$ 
  \item $\forall i > 0:\ \mathcal{S}_i$ is $\mathcal{N}$-linked
  \item conditions \ref{cond:I} and \ref{cond:II} hold:
 \begin{itemize}
  \item $cur(\mathcal{S}_i)<\gamma\quad \forall i>0$
  \item $d(\mathcal{S}_i,\mathcal{S}_j)>\delta\cdot \max\{s(\mathcal{S}_i),s(\mathcal{S}_j)\}\ \forall i,j>0, i\neq j$.
 \end{itemize}
 \end{enumerate}
The set of feasible sets of clusters is denoted by $feas_{\mathcal{N}}(\mathcal{S})$.
\end{defi}
The clusters $\mathcal{S}_i,\ i> 0$ identify the baselines, $\mathcal{S}_0$ constitutes the clutter cluster containing SPs not belonging to any baseline.
We identify the baseline corresponding to $\mathcal{S}_i$ with the polygonal chain of the projected SPs $\mathcal{S}_i^c$ which follow the regression curve $c_{\mathcal{S}_i}(t)$, see Fig.~\ref{fig:fin}.
The number $L\in\mathbb{N}$ of baselines is (a-priori) unknown.  
In the following, we will incorporate domain knowledge to promote SPs belonging to different baselines not to be $\mathcal{N}$-linked. Hence, clusterings with erroneously connected baselines are not feasible anymore. This is done by a modification of the neighborhood system $\mathcal{N}$.

Since baselines of different text orientations should not contribute to the same cluster, we adjust the initial neighborhood system $\mathcal{N}$ by removing edges $\boldsymbol{e}_{\boldsymbol{p},\boldsymbol{q}}$ 
of SPs with substantially different local orientations: $\left|\theta_{\boldsymbol{p}}-\theta_{\boldsymbol{q}}\right|\mod\pi > \frac{\pi}{4}$. In addition, it is an ease to incorporate layout 
information by further adjusting $\mathcal{N}$. 
The layout information encoded by the separator image $S$ (Fig.~\ref{fig:aruout}) can be incorporated by taking into account the connectivity of SPs in $S$. All edges $\boldsymbol{e}_{\boldsymbol{p},\boldsymbol{q}}\in\mathcal{N}$ for which a separator is crossed, i.e., $\Gamma(\boldsymbol{e}_{\boldsymbol{p},\boldsymbol{q}},S)>\eta$ or $\max_{\tau}S(g(\tau;\boldsymbol{e}_{\boldsymbol{p},\boldsymbol{q}}))> 2\cdot\eta$ ($g$ of Def.~\ref{def:cof}) holds, are removed, see Fig.~\ref{fig:finB}.

Finally, a common scenario is the baseline detection with given text regions. We assume that the text regions are represented by closed polygonal chains $\boldsymbol{R}_1,...,\boldsymbol{R}_N$. This additional layout information (if available) is easy to integrate.
All edges for which 
$\nexists \boldsymbol{R}_i:\ \boldsymbol{R}_i\text{ contains } \boldsymbol{p},\boldsymbol{q}$
holds are removed.
Roughly speaking, a closed polygonal chain \textit{contains} an SP if for all "ways" from the SP to the image border one have to cross the polygonal chain.
Hence, SPs which are part of different non-overlapping text regions are not $\mathcal{N}$-linked any more. 
Thus, each baseline $\mathcal{S}_i,\ i>0$ is entirely contained in one text region for all feasible sets. The resulting neighborhood system is still denoted by $\mathcal{N}$.
\begin{rem}
\label{rem:fe}
 For all experiments, we have chosen fixed values of $\gamma= 0.3$, $\delta = 0.5$ (Def.~\ref{def:fe}) and $\eta=0.125$.
\end{rem}
After reducing the neighborhood system, we now introduce the total baseline energy. 
We will assign an energy to all feasible sets and aim for an optimal one. 
This allows for the formulation of the clustering problem to be solved. 
\begin{defi}[total baseline energy]
 Let $B$ be a baseline image, $\mathcal{N}$ a neighborhood system and $\mathscr{P}=\left\{\mathcal{S}_0, ... , \mathcal{S}_L \right\}$ a set of clusters over $\mathcal{S}$. 
 With $\mathcal{N}(\mathcal{S}_i)=\{\boldsymbol{e}_{\boldsymbol{p},\boldsymbol{q}}\in\mathcal{N}\ |\ \boldsymbol{p},\boldsymbol{q}\in\mathcal{S}_i\}\subset\mathcal{N}$
 the \textit{total baseline energy} is defined by
 \begin{align*}
  b(\mathscr{P})=\sum_{i=1}^L\ \sum_{\boldsymbol{e}_{\boldsymbol{p},\boldsymbol{q}}\in\mathcal{N}(\mathcal{S}_i)}\Gamma(\boldsymbol{e}_{\boldsymbol{p},\boldsymbol{q}},B).
 \end{align*}
\end{defi}
Finally, the clustering problem can be formulated as 
\begin{align*}
 \mathscr{P^*}=\argmax_{\mathscr{P}\in feas_{\mathcal{N}}(\mathcal{S})}b(\mathscr{P}).
\end{align*}
Because there could be a huge number of feasible sets of clusters for large $\mathcal{S}$, we introduce a greedy clustering algorithm.
The proposed algorithm clusters edges of $\mathcal{N}$ instead of clustering SPs.
If an edge is assigned to a cluster (set) of edges, we assign both corresponding SPs to the corresponding cluster of SPs. 
In a first step, the set of edges in $\mathcal{N}$ is sorted in decreasing order w.r.t.
\begin{align*}
 \left(1- \frac{\left\| \boldsymbol{p}-\boldsymbol{q}\right\|_{\theta(\{\boldsymbol{p},\boldsymbol{q}\})}}{\left\|\boldsymbol{p}-\boldsymbol{q}\right\|_2} \right)\cdot \Gamma(\boldsymbol{e}_{\boldsymbol{p},\boldsymbol{q}},B).
\end{align*}
Hence, the sorting takes into account the $B$-connectivity value of an edge and discounts it if $\boldsymbol{e}_{\boldsymbol{p},\boldsymbol{q}}$ is rather orthogonal to $\theta(\{\boldsymbol{p},\boldsymbol{q}\})$. 
Discounted edges are less likely part of a baseline and are therefore sorted to the end of the list. The sorted list is denoted by $\boldsymbol{N}$. 
This avoids that these edges are falsely assigned to baseline clusters which are composed of just a few correct edges (statistics of the cluster are not reliable, yet). 
Given $\mathcal{S}$ and $\boldsymbol{N}$, the proposed clustering algorithm assigns all edges to the clutter cluster ($\mathcal{S}_0$). It iteratively moves edges to baseline clusters such that the resulting set of clusters remains feasible and the total baseline energy increases. The algorithm is shown in detail in the supplements (Alg.~S.2). 
\section{Experiments}
\label{sec:exp}
The experiment section is divided into $4$ subsections. First, we investigate the influence of the training set size as well as the influence of different data augmentation strategies.
This is followed by an investigation of the performance of the proposed method if it is applied to images with curved or arbitrarily oriented text lines.
The third subsection presents and compares results of different versions of our proposed NPL architectures on the very heterogeneous and challenging cBAD dataset \cite{gruning2017read,diem_markus_2017_257972}.
We perform statistical tests to show the statistical significance of the stated conclusion -- the superiority of the proposed ARU-Net in a two-stage workflow over other architectures and a single-stage workflow. 
Finally, we compare the proposed method against other state-of-the-art methods on the datasets of $3$ recently hosted competitions. 
As mentioned in Sec.~\ref{sec:met} we will follow the similarity score of \cite{gruning2017read} (F-value) to measure the quality of the baseline detection.
The configuration for all experiments including the hyperparameters of the network architecture as well as the training are summarized in Tab.~\ref{tab:hyp}. This configuration is the result of an extensive search in the hyperparameter space and results in impressive results for various scenarios/datasets.
\begin{table}[ht]
\renewcommand{\arraystretch}{1.3}
\centering
\caption{\textbf{Hyperparameters --} The architecture and training configuration which were used in this work are described.}
\begin{threeparttable}
\label{tab:hyp}
\begin{tabular}{|p{0.475\textwidth-2\tabcolsep-2\fboxrule}|}
\hline
\textbf{Image pre-processing}: input image $I$ is downscaled by a factor of $2$ for $\max\{I_h,I_w\}< 2000$, $3$ for $2000\leq\max\{I_h,I_w\}< 4800$ or $4$ followed by a normalization to mean $0$ and variance $1$ (on pixel intensity level)\\
 \hline
\textbf{RU-Net architecture, see Fig.~\ref{fig:unet} \& \ref{fig:res}}: number of scale spaces: $6$, initial feature depth: $8$, residual depth (activated layers in a residual block): $3$, feature increasing and spatial decreasing factor: $2$, activation function: ReLu, kernel size: $3\times 3$, stride: $1$\\
\hline
\textbf{A-Net architecture}: $4$ layer CNN, activation function: ReLu, kernel size: $4\times 4$, stride: $1$, maxpooling of size $2\times 2$ after each convolution, feature number: $12,16,32,1$\\
\hline
\textbf{ARU-Net architecture, see Fig.~\ref{fig:arunet}}: number of image scales: $5$, classifier: $4\times 4$ convolution layer with softmax activation\\
\hline
\textbf{Training}: weight initialization: Xavier, optimizer: RMSprop, learning rate: $0.001$, learning rate decay per epoch: $0.985$, weight decay on the $L_2$ norm: $0.0005$, exponential moving average on the model weights: $0.9995$, mini batch size: $1$ (due to memory limitations of the GPU), early stopping: none (trained for a fixed number of epochs)\\
\hline
\end{tabular}
\end{threeparttable}
\end{table}

Since no early stopping based on the loss for any validation set is used, we train on the entire training set.
The ARU-Net workflow for training and inference (Tensorflow code) as well as a trained network are freely available\footnote{https://github.com/TobiasGruening/ARU-Net}. The ARU-Net training takes \SIrange{3}{24}{h} from scratch (dependent on the number of epochs and samples per epoch) 
on a Titan X GPU. The inference time per image ranges from \SIrange{2}{12}{s} per image on a dual core laptop (Intel Core i7-6600U with 16GiB RAM), this reduces to \SIrange{0.5}{2}{s} running the ARU-Net on the Titan X.  
\subsection{Influence of Training Sample Number and Data Augmentation}
A major drawback of state-of-the-art approaches (Sec.~\ref{sec:rel}) is the need for an extensive expert tuning if confronted with scenarios which are not already covered. 
But the eligibility for an usage at industrial scale depends on the possibility to easily adapt at reasonable cost. For approaches relying on machine learning, this reduces to two questions:
\begin{itemize}
 \item What about the amount of ground truth needed? 
 \item What about the effort of ground truth production?
\end{itemize}
Concerning the second question, we refer to the automatic generation of pixel ground truth given the baseline ground truth. The annotation of (polygonal) baselines for a document image is quite easy and does not need remarkable expert knowledge compared to, e.g., ground truth production for ATR systems for 
historical handwritings or even the text line annotation at surrounding polygon level. The effort is reduced to several minutes per page by using platforms such as Transkribus\footnote{https://transkribus.eu}. In the following, we want to examine the first question.

The influence of training dataset size along with different data augmentation strategies is investigated for the freely available Bozen dataset\footnote{https://zenodo.org/record/218236} \cite{bozen}, see Fig.~S.1 (supplements). 
This dataset is a subset of documents from the Ratsprotokolle 
collection of Bozen composed of minutes of the council meetings held from 1470 to 1805 and consists of 400 pages. It is written in Early Modern German. Baseline ground truth information is available in form of PAGE\footnote{http://www.primaresearch.org/tools
} XML. The dataset is quite challenging concerning layout analysis issues. Most of the pages consist of a single main text region with many difficulties for line detection and extraction, e.g., bleed through, touching text lines and marginalia.
For the following experiments, we have randomly divided the Bozen set in a set of training samples $\mathcal{T}$ of size 350 and a test set of size 50. 
In a first step, we randomly set up a chain of subsets of $\mathcal{T}$
\begin{align*}
 \mathcal{T}_1\subset\mathcal{T}_3\subset\mathcal{T}_5\subset\mathcal{T}_{10}\subset\mathcal{T}_{30}\subset\mathcal{T}_{50}\subset\mathcal{T}_{100}\subset\mathcal{T}_{200}\subset\mathcal{T}_{350},
\end{align*}
where $\mathcal{T}_i$ contains $i$ training samples (pages and pixel ground truth). Since we expect an influence of the choice of training samples (= sorting of $\mathcal{T}$), we repeat the mentioned procedure $4$ times. Notably, the test set remains untouched. Finally, we got $45$ training sets -- five of each quantity. For each set, we trained the RU-Net for $100$ epochs with $256$ images per epoch. Therefore, we randomly choose samples of the training set and remove them from the set. If each element of the training set was used for training once, we start again with the initial training set. Hence, it does not matter whether the number of training samples per epoch exceeds the size of the training set or not. 
This procedure guarantees the same amount of training samples shown to the networks in training independent of the size of the training set.
The RU-Net was chosen instead of the ARU-Net, because of the homogeneity of the Bozen dataset concerning font size and resolution.
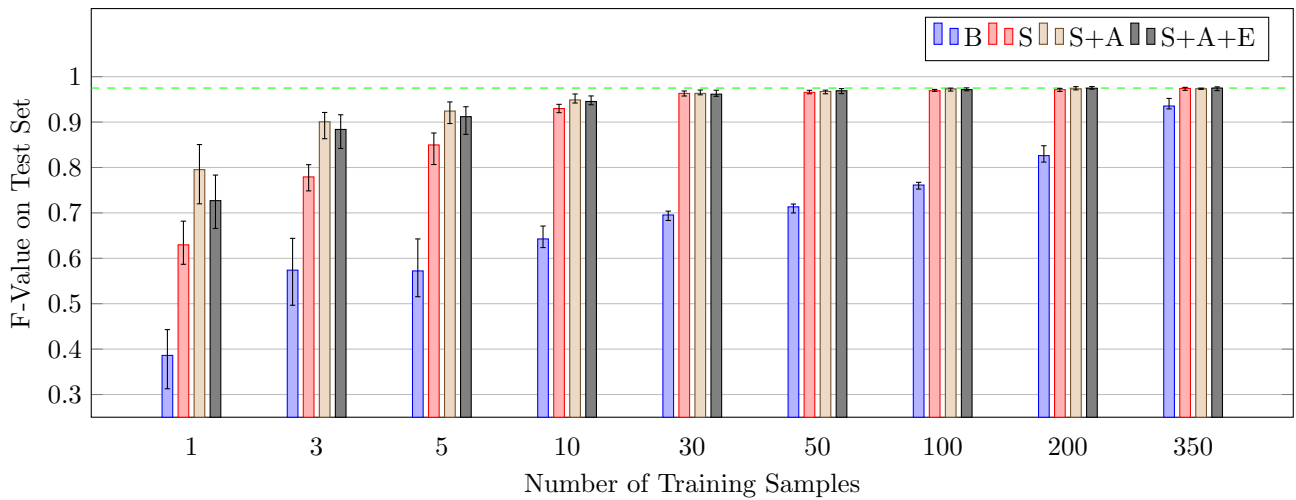
\begin{figure*}
	\begin{tikzpicture}
	\begin{axis}[
	width  = \textwidth,
	xlabel=Number of Training Samples,
	ylabel=F-Value on Test Set,
	height = 7cm,
	major x tick style = transparent,
	ybar,
	bar width=4pt,
	ymajorgrids = true,
	symbolic x coords={1,3,5,10,30,50,100,200,350},
	xtick = data,
	ytick={0.3,0.4,0.5,0.6,0.7,0.8,0.9,1.0},
	ymin=0.25,
	ymax=1.15,
	legend cell align=left,
	legend style={legend columns=-1},
	]
	\draw [green, style=dashed] ({rel axis cs:0,0}|-{axis cs:350,0.97492}) -- ({rel axis cs:1,0}|-{axis cs:350,0.97492});
	\addplot+[error bars/.cd, y dir=both, y explicit, error bar style={black},error mark options={rotate=90,mark size=1pt}] table [y index=1, y error plus index=2, y error minus index=3] {bozen_res.txt};
	\addplot+[error bars/.cd, y dir=both, y explicit, error bar style={black},error mark options={rotate=90,mark size=1pt}] table [y index=1, y error plus index=2, y error minus index=3] {bozen_res_s.txt};
	\addplot+[error bars/.cd, y dir=both, y explicit, error bar style={black},error mark options={rotate=90,mark size=1pt}] table [y index=1, y error plus index=2, y error minus index=3] {bozen_res_s_a.txt};
	\addplot+[error bars/.cd, y dir=both, y explicit, error bar style={black},error mark options={rotate=90,mark size=1pt}] table [y index=1, y error plus index=2, y error minus index=3] {bozen_res_s_a_e.txt};
	\legend{ B, S, S+A, S+A+E}
	\end{axis}
	\end{tikzpicture}
	\caption{\textbf{Influence of the number of training samples and of different data augmentation strategies --} The bar height represents the mean F-value. The error bars encode min-max values of the $5$ experiments (not the standard deviation). The dashed green line marks 
		the maximum mean value of $0.975$ achieved for $350$ trainings samples. For a description of the different augemantation strategies: B, S, S+A and S+A+E, see main text.}
	\label{fig:tS}
\end{figure*}
We trained the RU-Net from scratch on all $45$ sets in $4$ different scenarios. For training purposes the image pre-processing mentioned in Tab.~\ref{tab:hyp} is disabled. Instead, the training samples $(I,\mathcal{G}_I)_i$ are pre-processed following one of the four strategies:
\begin{enumerate}
 \item subsampled by a constant factor of $3$ (no further data augmentation - one training sample per element of the training set) -- B
 \item randomly subsampled by a factor $s\in\left[2,5\right]$ -- S
 \item S + random affine transformation (three corner points of the image are randomly shifted within a circle of diameter $0.025\cdot\max (\mathcal{I}_h,\mathcal{I}_w)$ around there original position) -- S + A
 \item S + A + elastic transformation \cite{Simard2003} -- S + A + E
\end{enumerate}
For the test set the images were sub-sampled by the constant factor of $3$ in all scenarios.
The results of these $180$ experiments are shown in Fig.~\ref{fig:tS}.
One can see that all $3$ data augmentation strategies significantly improve the performance compared to the base (B) strategy. Notably, for small numbers of training samples 
the min-max difference is much larger than for higher number of training samples. Hence, if just a few training samples are available, the choice of these is of importance. 
The best mean F-value ($0.975$) is achieved for all $350$ training samples with the S + A + E strategy. Nevertheless, there only is a negligible loss in performance for $200$ or $100$ training samples. Even for $30$ training samples,
a F-value of $0.963$ is achieved for the S+A strategy, which is sufficient for most applications, see Fig.~S.1 (supplements). 
This results in a quite acceptable effort for ground truth production making the presented approach interesting even for industrial production.
The S + A data augmentation strategy will be the default for the rest of this work.

Of course, the presented numbers are not directly transferable to collections with pages of entirely different scenarios, e.g., census tables mixed with postal cards mixed with ... .
One would expect that more than $30$ training samples are necessary for this kind of scenario.
Nevertheless, the presented experiment reflects a common situation: One has a robust baseline detector which was trained on very heterogeneous data (see Sec.~\ref{ssec:cbad}), but this 
detector does not work satisfyingly well for a certain (in most cases quite homogeneous) collection. The numbers presented here give a hint concerning the effort of ground truth production necessary in this scenario.
\subsection{Curved and Oriented Text Lines}
In this subsection, we demonstrate the ability of the introduced approach to handle curved or arbitrarily oriented text lines. 
In a first experiment, the test set of the Bozen dataset was deformed to contain arbitrarily curved text lines. 
For this purpose, we utilized trigonometric functions with random period to simulate curved text lines in the test phase.
The RU-Net was trained ($5$ times) for $100$ epochs with $256$ samples per epoch on the Bozen training set using the S + A + E augmentation strategy with strong elastic deformations. 
We choose elastic transformations in training, because they simulate curves of different amplitudes and frequencies in the same image.
Furthermore, we increased the polynomial degree (Def.~\ref{def:rp}) to $5$ to enable the system
to handle the curvatures present in the test set. 
\begin{rem}
	Different methods were used to deform the images during training and test phases. 
	Hence, the system had to learn the concept of curved text lines instead of an inversion of the image degradation method used in the training phase.
\end{rem}
In a second experiment, we have trained an RU-Net ($5$ times) on arbitrarily oriented samples of the 
Bozen training set and evaluated the resulting networks on oriented pages of the test set. The results are shown in Tab.~\ref{tab:brc} and a few sample images are shown in Fig.~S.2 (supplements). 
For the curved scenario the results are as good as for the base scenario. In case of the oriented scenario the results are slightly worse, but still excellent. 
This demonstrates the applicability for images with curved or oriented text lines without remarkable adaptation of the workflow. Finally, we have trained five models with all degradations (affine, elastic, rotation) and evaluated this model on the three different scenarios. The corresponding F-values are depicted in Tab.~\ref{tab:brc}. The system is worse than the experts for the base and curved scenarios, but for the oriented scenario it even benefits from the additional elastic transformations. 
\begin{table}[ht]
\renewcommand{\arraystretch}{1.3}
\centering
\caption{\textbf{Results for the Bozen test set --} The results in the Base (B), Curved (C) and Oriented (O) scenario are depicted. The P-, R- and F-values are strongly related to the well known precision and recall measures, see \cite{gruning2017read}. Finally, the F-val for a single system trained with all degradations are shown for the different test lists (A).}
\begin{threeparttable}
\label{tab:brc}
\centering
\begin{tabular}{|c||c|c|c|}
\hline
& \textbf{$\varnothing$ P-val}& \textbf{$\varnothing$ R-val}& \textbf{$\varnothing$ F-val} [min,max]\\
 \hline
B& $\boldsymbol{0.977}$&  $\boldsymbol{0.973}$&$\boldsymbol{0.975}\ \left[0.969,0.977\right]$ \\
C & $\boldsymbol{0.980}$&  $\boldsymbol{0.969}$&$\boldsymbol{0.975}\ \left[0.973,0.976\right]$ \\
O & $\boldsymbol{0.963}$&  $\boldsymbol{0.966}$&$\boldsymbol{0.964}\ \left[0.958,0.967\right]$ \\
\hline\hline
 &  \textbf{F-val} (B)&   \textbf{F-val} (C)& \textbf{F-val} (O) \\
 \hline
A  & $\boldsymbol{0.953}$ & $\boldsymbol{0.957}$  & $\boldsymbol{0.968}$ \\
\hline
\end{tabular}
\end{threeparttable}
\end{table}
\subsection{U-Net vs. ARU-Net vs. Single-Stage Workflow}
\begin{table*}
	\renewcommand{\arraystretch}{1.3}
	\centering
	\caption{\textbf{Results for the cBAD test set --} The results for different neural network architectures and the workflow without Stage II (for the ARU-Net) are shown. 
		Each architecture is trained $5$ times on the cBAD train set. The results are sorted with respect to computational effort. 
		The last two columns indicate whether an architecture is superior to all before mentioned ones in terms of disjunct confidence intervals and the Tukey-Duckworth test.}
	\begin{threeparttable}
		\label{tab:cB}
		\centering
		\begin{tabular}{|c||c|c|c|c|}
			\hline
			\multirow{2}{*}{}& \multicolumn{2}{c|}{\textbf{$\varnothing$ F-val} [$95\%$ CI]}& \multirow{2}{*}{CI}& \multirow{2}{*}{T-D}\\
			& Simple Track & Complex Track &  & \\
			\hline
			ARU I\textdagger& $\boldsymbol{0.9627}\ \left[0.9615,0.9636\right]$ &$\boldsymbol{0.9081}\ \left[0.9071,0.9095\right]$ & &\\
			U & $\boldsymbol{0.9714}\ \left[0.9701,0.9721\right]$&  $\boldsymbol{0.9114}\ \left[0.9107,0.9122\right]$& \cmark & \cmark \\
			RU& $\boldsymbol{0.9756}\ \left[0.9744,0.9766\right]$& $\boldsymbol{0.9182}\ \left[0.9165,0.9203\right]$ & \cmark & \cmark \\
			ARU& $\boldsymbol{0.9781}\ \left[0.9772,0.9789\right]$ &$\boldsymbol{0.9223}\ \left[0.9214,0.9230\right]$ & \cmark & \cmark \\
			LARU& $\boldsymbol{0.9772}\ \left[0.9765,0.9780\right]$ &$\boldsymbol{0.9233}\ \left[0.9217,0.9249\right]$ & \xmark & \xmark \\
			\hline
		\end{tabular}
		\begin{tablenotes}
			\item[\textdagger] single-stage workflow -- baseline estimation by basic image processing methods (binarization of $B$ followed by a CC analysis, no usage of $S$)
		\end{tablenotes}
	\end{threeparttable}
\end{table*}
In Sec.~\ref{sec:met}, we have introduced the ARU-Net in a two-stage workflow. In this section, we will investigate its superiority over the classical U-Net as well as over a ''single-stage'' workflow. For this purpose we have trained the U-, RU-, ARU- and LARU-Net (each $5$ times -- random weight initialization and random training sample order) on the recently 
introduced cBAD dataset\footnote{https://zenodo.org/record/257972} \cite{diem_markus_2017_257972}. The LARU-Net is an ARU-Net with a separable MDLSTM\footnote{A separable MDLSTM layer is a concatenation of two (x- and y-direction) BLSTM layers} layer at the lowest resolution to incorporate full spatial context. 
The details of the dataset are described in \cite{gruning2017read}. In our opinion, this is the most challenging freely available dataset at the moment. We have trained each network for $250$ epochs, 
$1024$ training samples each epoch using the S + A data augmentation strategy. To assure the statistical significance of the posed superiority of the newly introduced architecture, we follow \cite{puig2017} and provide the results of a statistical analysis.
The choice of appropriate statistical tests is quite limited since we can't make any assumptions regarding the underlying distribution. 
We utilize $95\%$ confidence intervals (CI) provided by non-parametric bootstrapping \cite{Efron1987} as well as the Tukey-Duckworth test (level of significance: $5\%$) \cite{Tukey1959}.
The results obtained are summarized in Tab.~\ref{tab:cB}. The ARU-Net performs significantly (last two columns) better than all architectures with less computational effort. The LARU-Net could not prove its superiority and is therefore dismissed.
Furthermore, the results show that the introduction of the second stage is beneficial for the overall performance.
It has to be mentioned that the above comparison is not fair concerning the number of trainable parameters -- U - $2.16$, RU - $4.13$, ARU - $4.14$, LARU - $6.25$ (in millions) -- nor concerning the training or even inference time. The comparison is about different architectures which, theoretically, have different capabilities, and whether they make good use of them or not. For instance, the LARU-Net should be capable of incorporating a more detailed spatial context, but in fact it does not benefit (in our settings) from this capability.
\subsection{Comparison against the State of the Art}
In this subsection, we compare the proposed framework against the state of the art. We have chosen the $3$ most recent competitions on text line detection for historical documents, namely: ICDAR 2015 
competition on text line detection in historical documents \cite{Murdock2015}, ICDAR2017 Competition on Layout Analysis for Challenging Medieval Manuscripts (Task 2) \cite{foti2017} and cBAD: ICDAR2017 Competition on Baseline Detection \cite{diem2017}.
We will not further introduce the datasets or metrics used and refer to the competition papers. 
\subsubsection{ICDAR 2015 Competition on Text Line Detection in Historical Documents (ANDAR-TL)}
The ARU-Net was trained on the cBAD training set\footnote{The competition training data was not available to the authors.}. 
This competition aims at the origin point (OP) detection. An OP is roughly spoken the lower left "corner" of a text line.
Hence, we calculate the left most point of each detected baseline. This is the output of our system for this competition. 
The achieved results are shown in Tab.~\ref{tab:15}. 
\begin{table}[ht]
	\renewcommand{\arraystretch}{1.3}
	\centering
	\caption{\textbf{Origing Point (OP) detection results for the ANDAR-TL test set --} Results for the dataset of \cite{Murdock2015} are shown. \#COR means the number of correctly detected OPs, \#DF means the number of detection failures (no OP detected by the system), \#DM means the number of detection misses (detected OP far away from the ground truth OP) and \#FP means the number of false positives. }
	\begin{threeparttable}
		\label{tab:15}
		\begin{tabular}{|c||c|c|c|c|c|}
			\hline
			 & \#COR & \#DF & \#DM & \#FP &cost\\
			\hline
			UNIFR& $2578$ & $3022$ & $6456$ & $267$ & $19.00$ \\
			IA-2& $5655$ & $407$ & $6032$ & $102$ & $14.51$ \\
			A2iA-3\textdagger& $6523$ & $2490$ & $2263$ & $181$ & $13.20$\\
			SNU\cite{Ahn2017}& $7741$ & $948$ & $2700$ & $25$ & $9.77$\\
			\cite{gr2017}& $8015$ & $517$ & $2860$ & $21$ & $8.19$\\
			ours& $9610$ & $358$ & $1942$ & $83$ & \underline{$5.39$}\\
			\hline
		\end{tabular}
		\begin{tablenotes}
			\item[\textdagger] According to \cite{Eskenazi2017} this is an extension of \cite{Moysset2015}.
		\end{tablenotes}
	\end{threeparttable}
\end{table}
Since the ARU-Net was not trained on the original training data, it is hard to compare its results to the other ones. 
Nevertheless, we would like to stress the fact, that trained systems usually perform better if training set and test set are sampled from the same distribution.
E.g., the ARU-Net trained on the cBAD training set achieves an average F-value of $0.9605$ for the Bozen test set, which is worse than the F-vlaue of $0.9750$ of the system trained solely on the Bozen training set, see Tab.~\ref{tab:brc}.
This indicates (but does not prove) the superiority of the presented method over the other methods in Tab.~\ref{tab:15}.
\begin{figure*}[ht]
	\centering
	\includegraphics[width=0.31\textwidth]{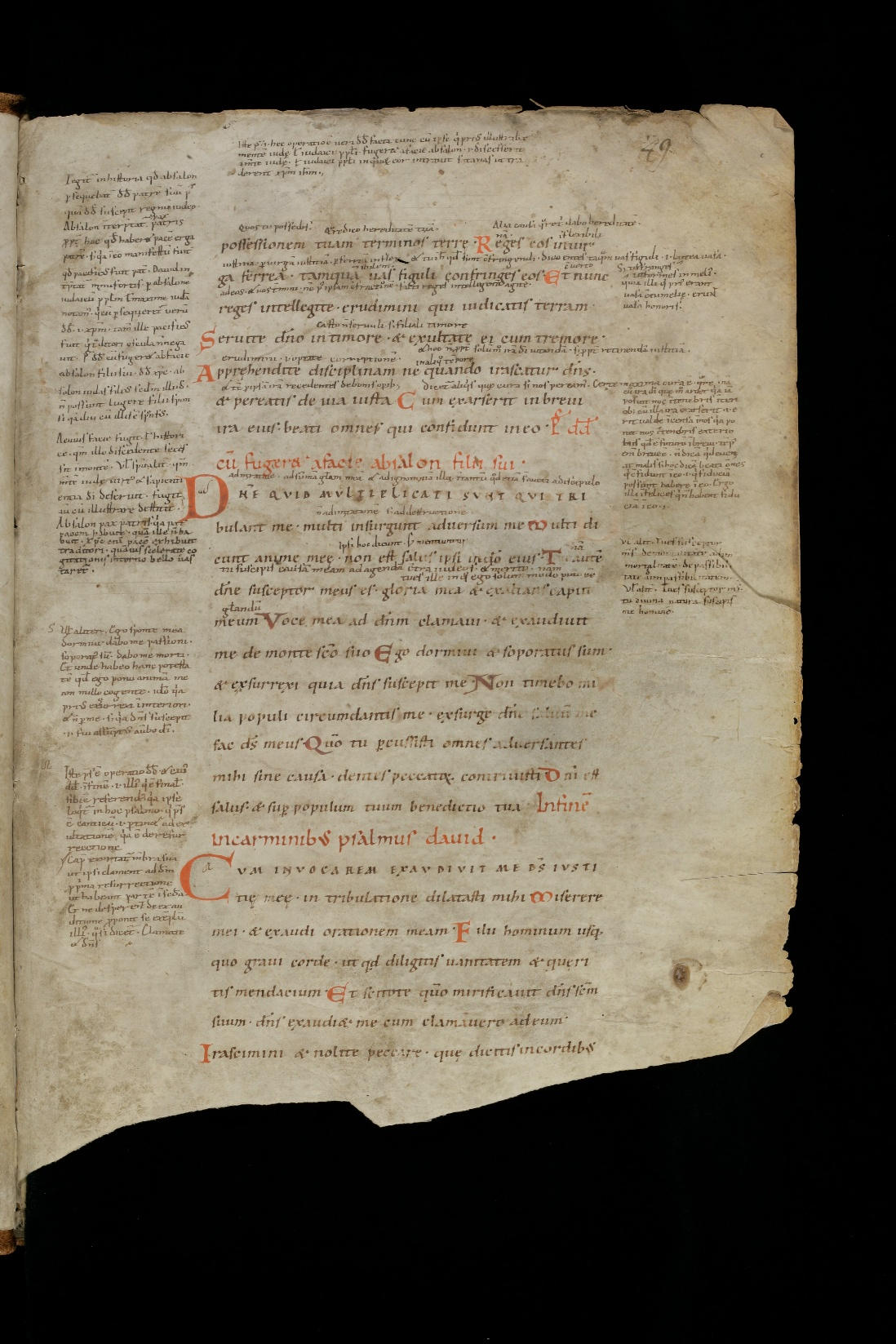}
	\includegraphics[width=0.31\textwidth]{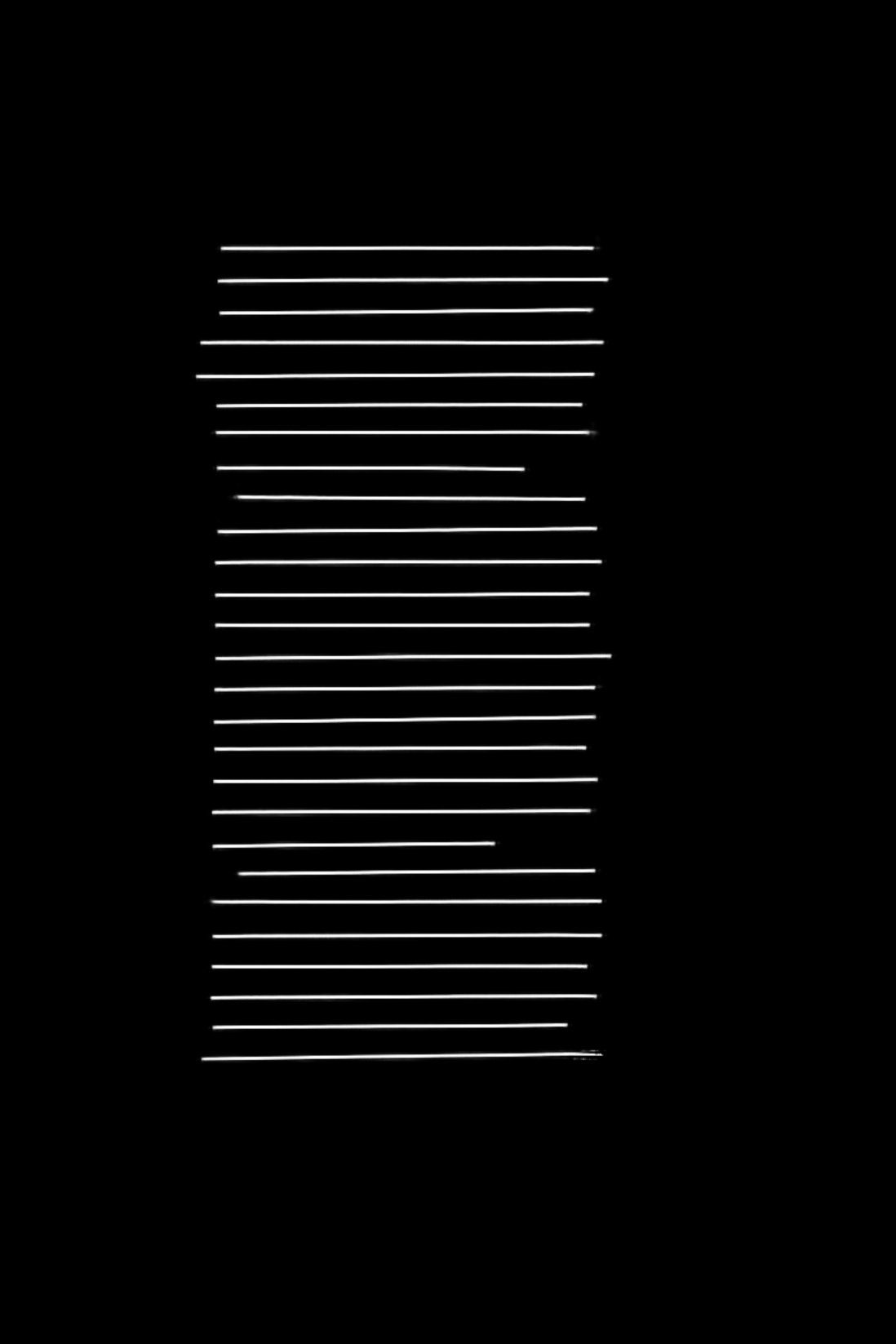}
	\includegraphics[width=0.31\textwidth]{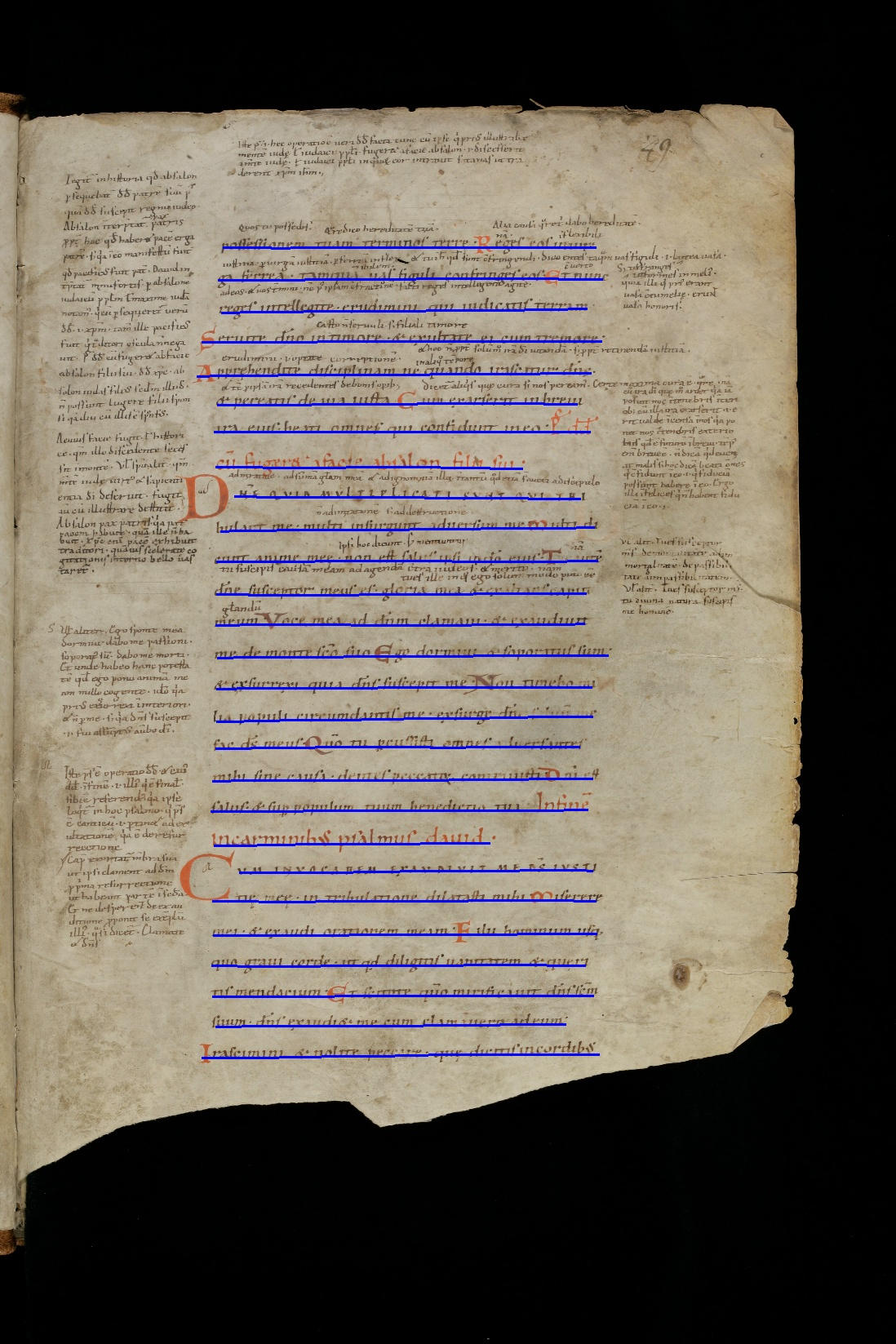}
	\caption{\textbf{Results for an image of the CSG18 subset of the test set --} The original image (only the main text lines were ground truthed), the baseline image generated by the trained ARU-Net and the baselines detected by the proposed method are shown (from left to right).}
	\label{fig:17a}
\end{figure*}
\subsubsection{ICDAR2017 Competition on Layout Analysis for Challenging Medieval Manuscripts (Task 2)}
The DIVA-HisDB dataset consists of $150$ annotated pages of three different medieval manuscripts with challenging layouts, see Fig.~\ref{fig:17a}.
The ARU-Net was trained for $250$ epochs $1024$ samples per epoch on the competition training data\footnote{http://diuf.unifr.ch/main/hisdoc/diva-hisdb} provided by the competition organizers. This allows an entirely fair comparison to the participant's results, see Tab.~\ref{tab:17a}. 
\begin{table}[ht]
\renewcommand{\arraystretch}{1.3}
\caption{\textbf{Results for the ICDAR2017 Competition on Layout Analysis for Challenging Medieval Manuscripts -- } The F-values for Task 2 of all participants and the proposed method are shown for the different subsets of the test set.}
\centering
\begin{threeparttable}
\label{tab:17a}
\begin{tabular}{|c||c|c|c|c|}
\hline
 & CB55 & CSG18 & CSG863 & overall \\
\hline
CVML& $0.9534$ & $0.8734$ & $0.9751$ & $0.9340$ \\
BYU& $0.9597$ & \underline{$0.9879$} & $0.9830$ & $0.9768$ \\
CITlab& $0.9896$ & $0.9853$ & $0.9716$ & $0.9822$\\
ours& \underline{$0.9980$} & $0.9828$ & \underline{$0.9889$} & \underline{$0.9899$}\\
\hline
\end{tabular}
\end{threeparttable}
\end{table}
The proposed method substantially outperforms the winning one
and reduces the error (the gap to $1.0$) by $43.26\%$ (relatively).  The specialty of this competition was, that the methods should focus on a special kind of text, e.g., comments were not annotated as text. Hence, the ARU-Net had to learn to distinguish
between different types of text. The output of the ARU-Net and the detected baselines for a sample image of the CSG18 subset of the test set are shown in Fig.~\ref{fig:17a}. One can see, that the ARU-Net entirely ignores 
all text entities not regarded (in this competition) as main text. Remarkably, no further information besides the input image is provided to the ARU-Net.
\subsubsection{cBAD: ICDAR2017 Competition on Baseline Detection}
\label{ssec:cbad}
We compare our average result for the ARU-Net (see Tab.~\ref{tab:cB}) to the results presented in \cite{diem2017}, see Tab.~\ref{tab:cbad}. 
\begin{table}[ht]
\renewcommand{\arraystretch}{1.3}
\centering
\caption{\textbf{Results for the cBAD test set --} The P-, R- and F-values of all participants and of the proposed method for the simple and complex track of the cBAD: ICDAR2017 Competition on Baseline Detection are shown.}
\begin{threeparttable}
\label{tab:cbad}
\centering
\begin{tabular}{|c||c|c|c|c|}
\hline
\multirow{2}{*}{}& \multicolumn{2}{c|}{Simple Track}& \multicolumn{2}{c|}{Complex Track}\\
 & P/R-val & F-val & P/R-val & F-val\\
 \hline
LITIS & $0.78/0.84$ & $0.807$& --/-- & --\\
\cite{aldavert2018manuscript} & $0.75/0.93$ & $0.827$& --/-- & --\\
UPVL & $0.94/0.86$ & $0.894$& $0.83/0.61$ & $0.702$\\
\cite{renton2018fully} & $0.88/0.88$ & $0.880$& $0.69/0.77$ & $0.730$\\
BYU & $0.88/0.91$ & $0.892$& $0.77/0.82$ & $0.796$\\
\cite{quiros2018multi}  & --/-- & -- & $0.85/0.85$ & $0.851$\\
\cite{fink2018baseline} & $0.97/0.97$ & $0.971$& $0.85/0.86$ & $0.859$\\
\cite{oliveira2018dhsegment} & $0.88/0.97$ & $0.920$& $0.79/0.95$ & $0.860$\\
ours & $\boldsymbol{0.98}/\boldsymbol{0.98}$ & \underline{$\boldsymbol{0.978}$}& $\boldsymbol{0.93}/\boldsymbol{0.92}$ & \underline{$\boldsymbol{0.922}$}\\
\hline
\end{tabular}
\end{threeparttable}
\end{table}
Our method performs considerably better in both tracks compared to all submissions. Especially, the increase in performance for the complex track is massive. Remarkably, the winning team uses an U-Net based system with task specific pre- and postprocessing. This indicates that the newly introduced concepts and parametrization, which are presented in this work, significantly improve the capability of the classical U-Net.
Some results on chosen images of the cBAD test set are shown in  Fig.~S.3-S.5 (supplements). Notably, no further information besides the input image (and the text region information in the simple track) is provided to the ARU-Net nor to the second stage of the workflow during inference.
\section{Conclusion}
\label{sec:conc}
In this work we presented a machine learning based method for text line detection in historical documents. The text lines are represented by their baselines. The problem and the proposed method were introduced thoroughly. 
The proposed ARU-Net, which is a universal pixel labeling approach, was trained to predict the baseline position and the beginning and end of each text line. This enables the system to handle documents with complex layouts, e.g., tables, marginalia, multi columns layouts.
We have shown that the system can be trained from scratch with manageably few training samples for a complex but homogeneous collection. Remarkably, ground truth production is quite cheap. A ground truth sample is just a page with annotated baselines, which can be done in a few minutes per page. Notably, this annotation process is possible without any expert knowledge. This is a big advantage compared to classical image processing based methods, which typically demand for expert knowledge in the adaptation phase.
Therefore, one can expect that an adaptation on collections, which are not covered by the neural network, is possible by a wide audience of users (not only computer scientists) with reasonable ground truthing effort. 
The applicability of the proposed method was shown for straight, curved and oriented text lines as well as for a combined scenario.
The superiority of the proposed ARU-Net in the two-stage workflow over the classical U-Net and over a simplified workflow was shown and statistically verified.
Finally, we showed that the proposed method substantially outperforms the previous state of the art. 
Nevertheless, as one can see in Fig.~S.3-S.5 (supplements) there are still errors made by the system, e.g., missed baselines (see Fig.~S.4 -- bottom right), segmentation errors (see Fig.~S.5 -- bottom left), false positives (see Fig.~S.3 -- top left) or problems with strongly degraded documents (see Fig.~S.4 -- top left). But these errors do not seem to follow a certain deterministic principle, which is not surprising for a method based on machine learning.
However, we plan to test newly introduced concepts like capsules, memory augmentation and deeply supervised networks to further improve the system's performance.

\begin{acknowledgements}
NVIDIA Corporation kindly donated a Titan X GPU used for this
research. This work was partially funded by the European Union’s
Horizon 2020 research and innovation programme under grant
agreement No 674943 (READ – Recognition and Enrichment
of Archival Documents). Finally, we would like to thank Udo Siewert for his valuable comments and suggestions.
\end{acknowledgements}
\singlespacing
\bibliography{lit}
\setcounter{figure}{0} 
\renewcommand{\thefigure}{S.\arabic{figure}}    
\setcounter{algocf}{0}
\renewcommand{\thealgocf}{S.\arabic{algocf}}  
\begin{algorithm*}
	\DontPrintSemicolon
	\SetKwComment{Comment}{$\rhd$ }{}
	\SetKwInOut{Input}{input}\SetKwInOut{Output}{output}\SetKwInOut{Return}{return}
	\vspace{1mm}
	\Input{image $I$, corresponding baseline ground truth $\mathcal{G}_I$}
	\Output{pixel ground truth $G_I$}
	$B,S,N \gets 0$\Comment*{of dimension $I_h\times I_w$}
	\For{$\boldsymbol{P}=\left(\boldsymbol{p}_1,...,\boldsymbol{p}_n\right)\in\mathcal{G}_I$}{
		$\theta \gets $ local text orientation of $\boldsymbol{P}$ \Comment*{see Def.~3.3.2} 
		$d \gets $ interline distance of $\boldsymbol{P}$ \Comment*{see Def.~3.3.3} 
		$\boldsymbol{P}_b \gets $ polygonal chain of length $d$ and orient. $\theta + 90\degree$ centered at $\boldsymbol{p}_1$\\ 
		$\boldsymbol{P}_e \gets $ polygonal chain of length $d$ and orient. $\theta + 90\degree$ centered at $\boldsymbol{p}_n$\\ 
		draw $\boldsymbol{P}_b$ and $\boldsymbol{P}_e$ in $S$\Comment*{draw: follow the chain and set pixel values to $1.0$}
		draw $\boldsymbol{P}$ in $B$ 
	}
	$E \gets $ $3\times 3$ matrix of ones\\
	$S \gets S\oplus E$\Comment*{$\oplus$ morphological dilation }  
	$B \gets B\oplus E \land \lnot S$\\
	$N \gets \lnot S \land \lnot B$\\
	\Return{$G_I \gets \left[B,S,N \right]$}
	\vspace{1mm}
	\caption{Pixel Ground Truth Generation}
	\label{alg:gt}
\end{algorithm*}
\begin{algorithm*}
	\DontPrintSemicolon
	\SetKwComment{Comment}{$\rhd$ }{}
	\SetKwInOut{Input}{input}\SetKwInOut{Output}{output}\SetKwInOut{Return}{return}
	\vspace{1mm}
	\Input{Set of SPs $\mathcal{S}$ and sorted list of edges $\boldsymbol{N}$}
	\Output{optimized partition $\mathscr{P}^*$}
	$\mathcal{S}_0 \gets \mathcal{S},\ \mathscr{P}^* \gets \{\mathcal{S}_0\},\ n\gets 0$\\
	\While{$\left|\boldsymbol{N}\right| \neq n$}{
		$n\gets \left|\boldsymbol{N}\right|$\\
		\For(\Comment*[f]{Four possible cases dependent on the SPs}){$\boldsymbol{e}_{\boldsymbol{p},\boldsymbol{q}}\in \boldsymbol{N}$}{
			\uIf(\Comment*[f]{Case 1: add edge to existing cluster!}){$\exists i>0:\ \boldsymbol{p},\boldsymbol{q}\in \mathcal{S}_i$}{
				$\boldsymbol{N} \gets \boldsymbol{N}\setminus \{\boldsymbol{e}_{\boldsymbol{p},\boldsymbol{q}}\}$
			}
			\uElseIf(\Comment*[f]{Case 2: create new cluster?}){$\boldsymbol{p},\boldsymbol{q}\in \mathcal{S}_0$}{
				\If(\Comment*[f]{$\gamma,\ \delta$ of Rem.~3.3.8}){$\left\| \boldsymbol{p}-\boldsymbol{q}\right\|_{\theta(\{\boldsymbol{p},\boldsymbol{q}\})}<\delta \cdot s(\{\boldsymbol{p},\boldsymbol{q}\})$}{
					$\boldsymbol{N} \gets \boldsymbol{N}\setminus \{\boldsymbol{e}_{\boldsymbol{p},\boldsymbol{q}}\}$, $\mathcal{S}_{\left|\mathscr{P}^*\right|}\gets\{\boldsymbol{p},\boldsymbol{q}\}$, $\mathcal{S}_0\gets \mathcal{S}_0\setminus\mathcal{S}_{\left|\mathscr{P}^*\right|}$\\
					$\mathscr{P}^*\gets\mathscr{P}^*\cup\{\mathcal{S}_{\left|\mathscr{P}^*\right|}\}$ 
				}
			}
			\uElseIf(\Comment*[f]{Case 3: extend cluster $\mathcal{S}_i$?}){w.l.o.g. $\boldsymbol{p}\in\mathcal{S}_0\land \exists i>0:\ \boldsymbol{q}\in \mathcal{S}_i$}{
				\If{$cur(\mathcal{S}_i\cup \{\boldsymbol{p}\})<\gamma\ \land\ d(\mathcal{S}_i,\{\boldsymbol{p}\})<\delta\cdot s(\mathcal{S}_i)$}{
					\If{$d(\mathcal{S}_i\cup \{\boldsymbol{p}\},\mathcal{S}_j)>\delta\cdot s(\mathcal{S}_j)\ \forall j\neq i,\ j>0$}{
						$\boldsymbol{N} \gets \boldsymbol{N}\setminus \{\boldsymbol{e}_{\boldsymbol{p},\boldsymbol{q}}\}$, $\mathcal{S}_{i}\gets \mathcal{S}_{i}\cup \{\boldsymbol{p}\}$, $\mathcal{S}_0\gets \mathcal{S}_0\setminus\{\boldsymbol{p}\}$\\
					}
				}
			}
			\ElseIf(\Comment*[f]{Case 4: merge clusters $\mathcal{S}_i$, $\mathcal{S}_j$?}){$\exists i, j>0\ (i\neq j):\ \boldsymbol{p}\in\mathcal{S}_i\land \boldsymbol{q}\in \mathcal{S}_j$}{
				\If{$cur(\mathcal{S}_i\cup \mathcal{S}_j)<\gamma\ \land\ d(\mathcal{S}_i,\mathcal{S}_j)<\delta\cdot \min\left(s(\mathcal{S}_i),s(\mathcal{S}_j)\right)$}{
					$\boldsymbol{N} \gets \boldsymbol{N}\setminus \{\boldsymbol{e}_{\boldsymbol{p},\boldsymbol{q}}\}$, $\mathcal{S}_{i}\gets \mathcal{S}_{i}\cup \mathcal{S}_{j}$\\
					$\mathscr{P}^*\gets\mathscr{P}^*\setminus\{\mathcal{S}_j\}$
				}
			}
		}
	}
	\Return{$\mathscr{P}^*$}
	\vspace{1mm}
	\caption{SP Clustering}\label{alg:cl}
\end{algorithm*}
\begin{figure*}[ht]
	\centering
	\includegraphics[width=0.9\textwidth]{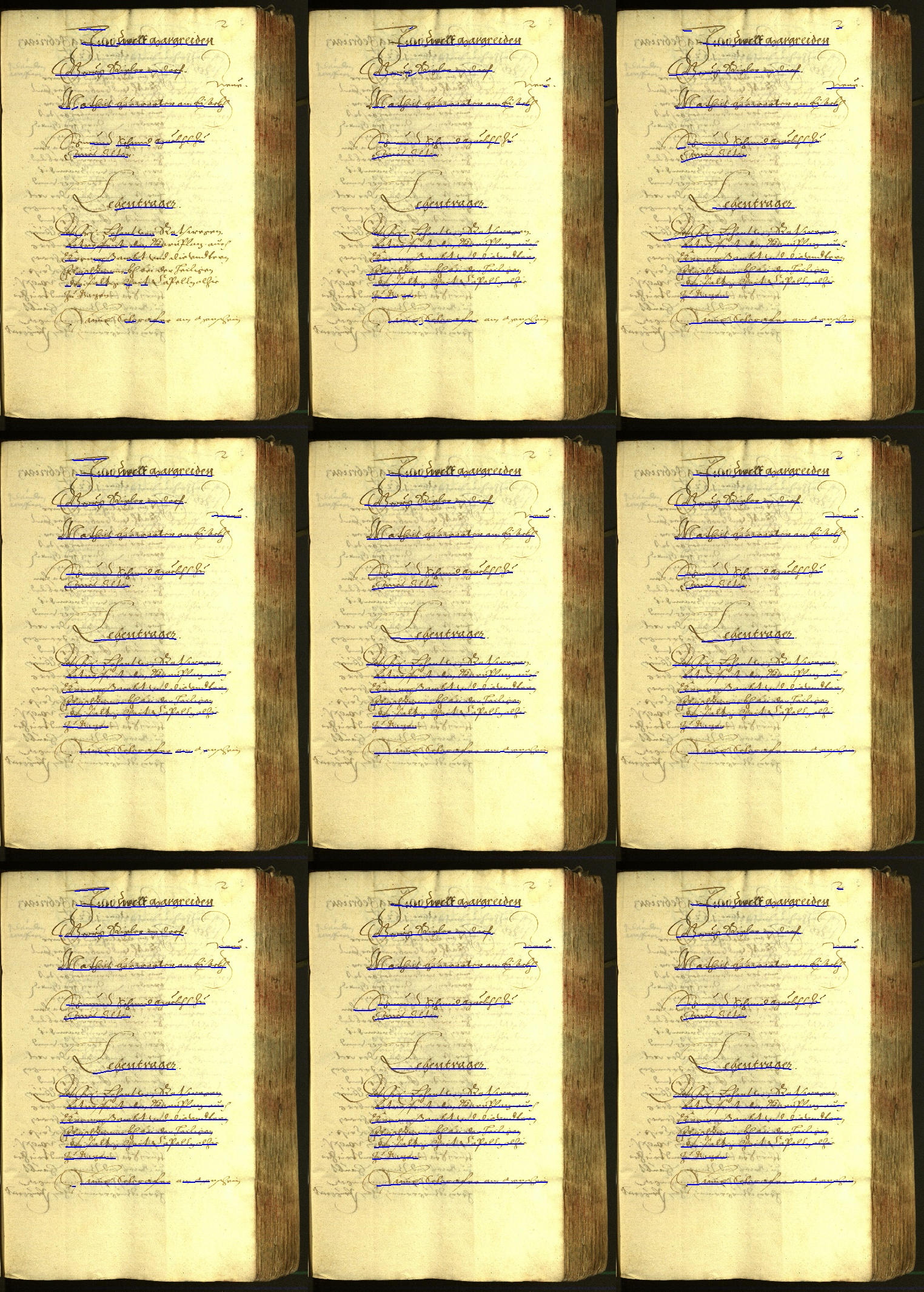}
	\caption{\textbf{Results for an image of the Bozen test set --} Results for RU-Nets trained on $5$, $30$ and $350$ training samples (left to right) with different data augmentation strategies B, S+A and S+A+E (top to bottom) are shown.}
	\label{fig:bDA}
\end{figure*}
\begin{figure*}
	\centering
	\includegraphics[width=0.9\textwidth]{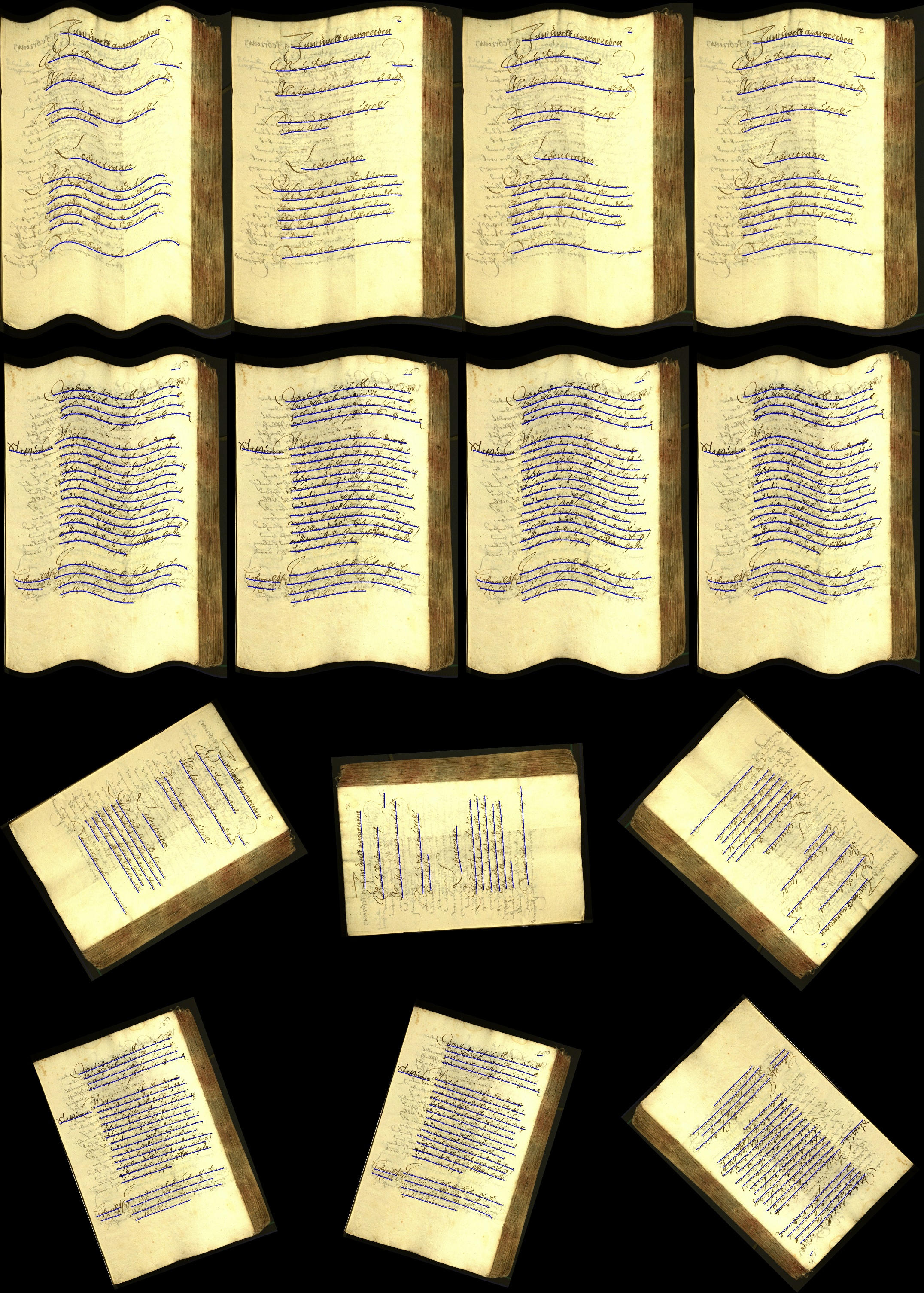}
	\caption{\textbf{Results for an image of the Bozen test set --} Results for two ``degraded'' images are shown. The images were arbitrarily curved and rotated.}
	\label{fig:brc}
\end{figure*}
\begin{figure*}
	\centering
	\includegraphics[width=\textwidth]{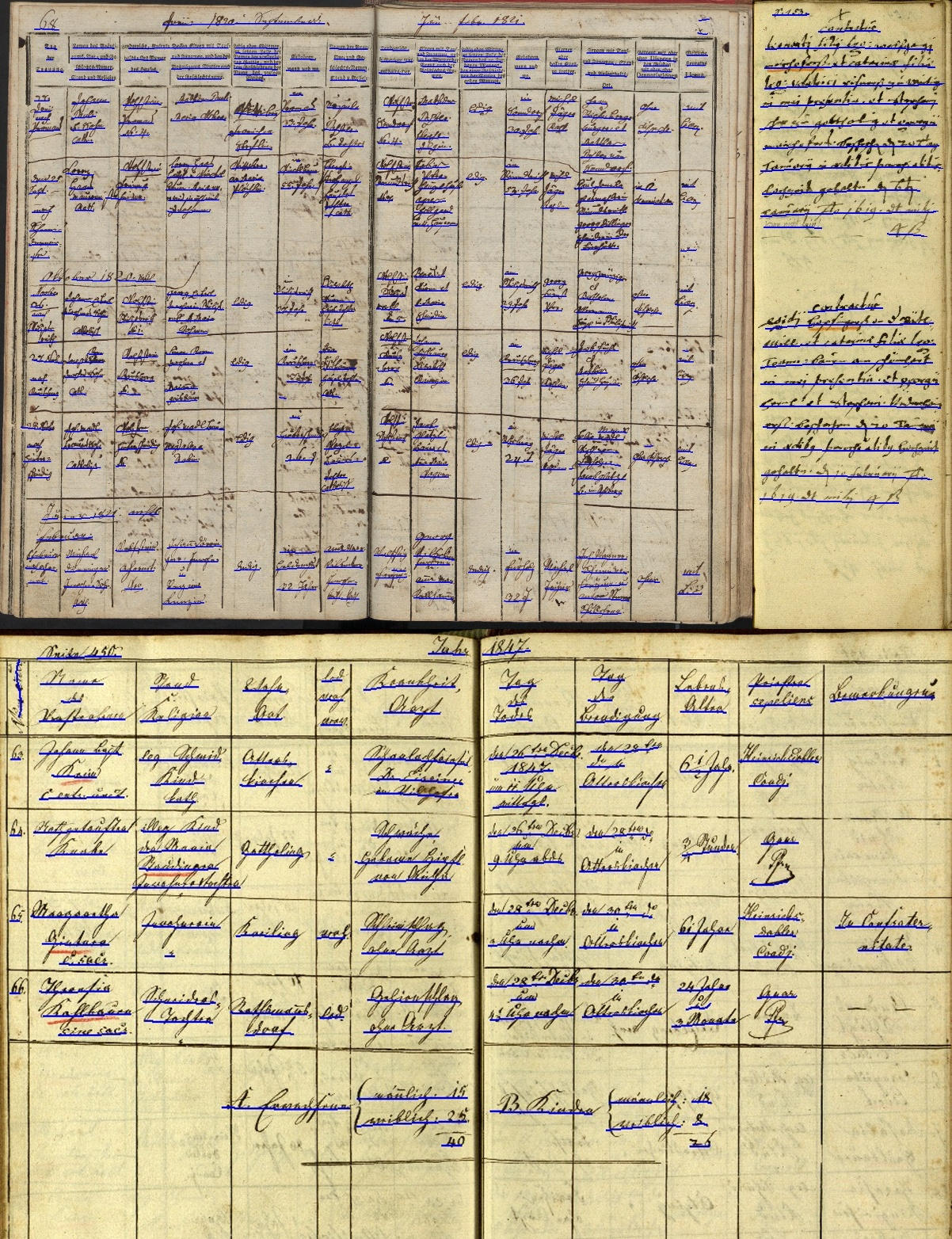}
	\caption{\textbf{Results for images of the cBAD test set --} Only the images without any layout information were used.}
	\label{fig:cbad1}
\end{figure*}
\begin{figure*}
	\centering
	\includegraphics[width=0.9\textwidth]{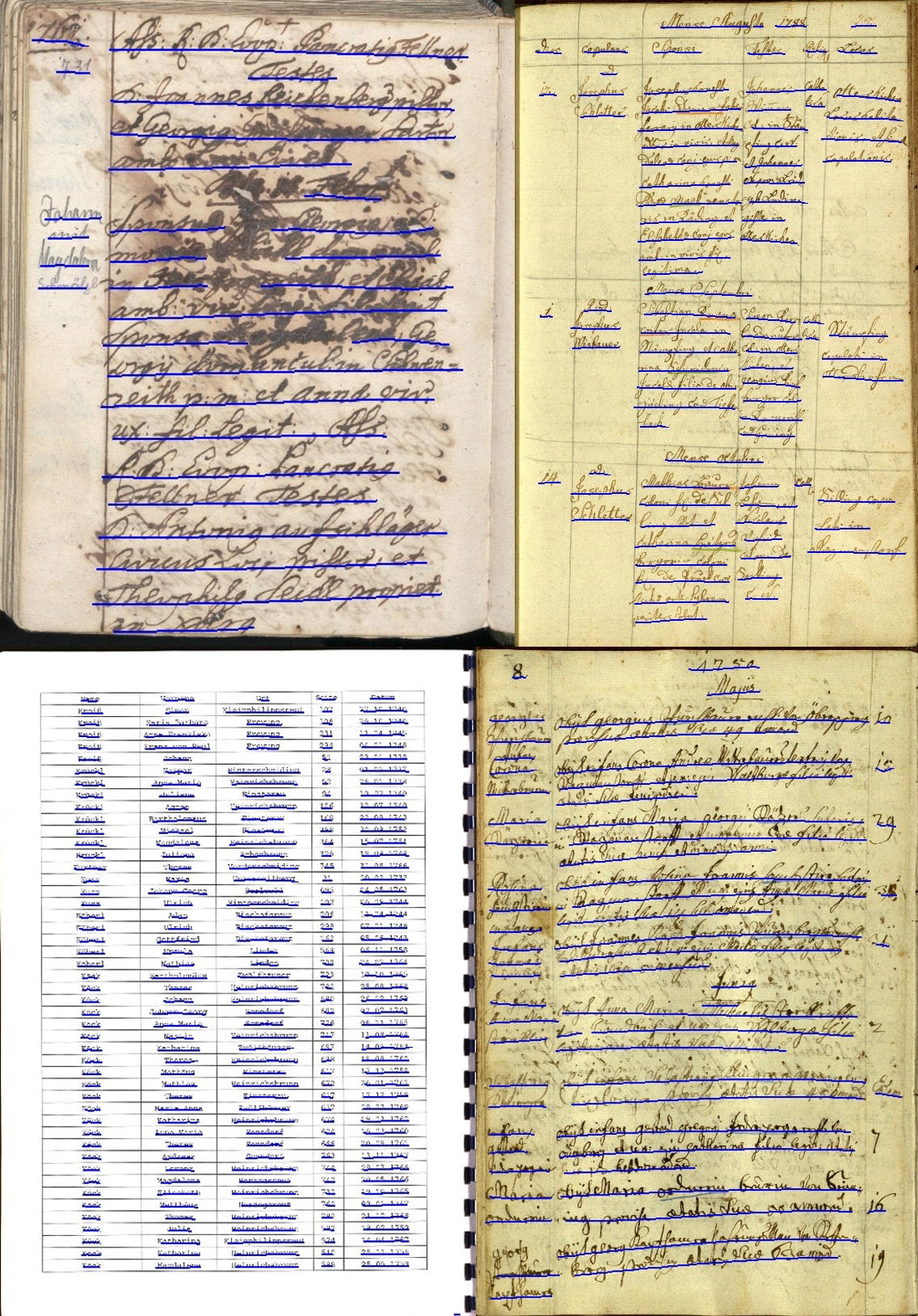}
	\caption{\textbf{Results for images of the cBAD test set --} Only the images without any layout information were used.}
	\label{fig:cbad2}
\end{figure*}
\begin{figure*}
	\centering
	\includegraphics[width=\textwidth]{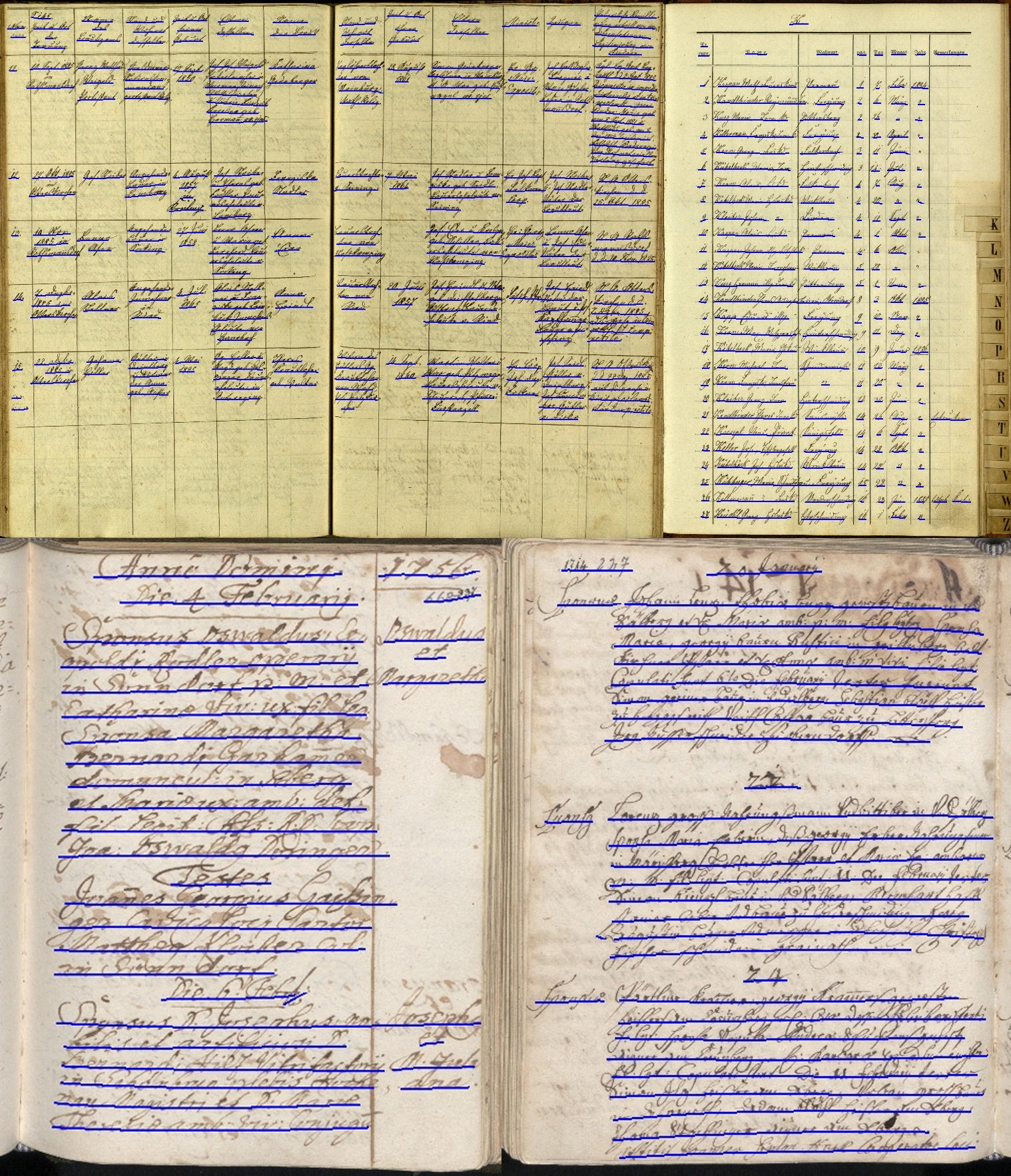}
	\caption{\textbf{Results for images of the cBAD test set --} Only the images without any layout information were used.}
	\label{fig:cbad3}
\end{figure*}
\end{document}